\documentclass[acmsmall,screen]{acmart}

\AtBeginDocument{%
  }

\setcopyright{acmlicensed}
\copyrightyear{2018}
\acmYear{2018}
\acmDOI{XXXXXXX.XXXXXXX}

\acmJournal{JACM}
\acmVolume{37}
\acmNumber{4}
\acmArticle{111}
\acmMonth{8}

\usepackage{array}

\usepackage{subcaption}

\usepackage{pifont}
\usepackage{booktabs}
\usepackage{multirow}
\usepackage[framemethod=TikZ]{mdframed}
\usepackage{hyperref}
\usepackage{bbm}
\usepackage{xcolor}
\usepackage{orcidlink}

\begin{document}

\title{M3-AGIQA: Multimodal, Multi-Round, Multi-Aspect AI-Generated Image Quality Assessment}

\author{Chuan Cui}
\authornote{Work is done during intership at Hefei High-Dimensional Data Technology Co.,Ltd.}
\email{cuichuan@tongji.edu.cn}
\affiliation{%
        \institution{Tongji University}
        \city{Shanghai}
        \country{China}
}

\author{Kejiang Chen}
\email{chenkj@ustc.edu.cn}
\affiliation{%
        \institution{University of Science and Technology of China}
        \city{Hefei}
        \country{China}
}

\author{Zhihua Wei}
\authornote{Corresponding author.}
\email{zhihua_wei@tongji.edu.cn}
\affiliation{%
        \institution{Tongji University}
        \city{Shanghai}
        \country{China}
}

\author{Wen Shen}
\email{wenshen@tongji.edu.cn}
\affiliation{%
        \institution{Tongji University}
        \city{Shanghai}
        \country{China}
}

\author{Weiming Zhang}
\email{zhangwm@ustc.edu.cn}
\affiliation{%
        \institution{University of Science and Technology of China}
        \city{Hefei}
        \country{China}
}

\author{Nenghai Yu}
\email{ynh@ustc.edu.cn}
\affiliation{%
        \institution{University of Science and Technology of China}
        \city{Hefei}
        \country{China}
}

\begin{abstract}

    The rapid advancement of AI-generated image (AIGI) models presents new challenges for evaluating image quality, particularly across three aspects: perceptual quality, prompt correspondence, and authenticity. 
    To address these challenges, we introduce M3-AGIQA, a comprehensive framework that leverages Multimodal Large Language Models (MLLMs) to enable more human-aligned, holistic evaluation of AI-generated images across both visual and textual domains. 
    Besides, our framework features a structured multi-round evaluation process, generating and analyzing intermediate image descriptions to provide deeper insight into these three aspects. 
    By aligning model outputs more closely with human judgment, M3-AGIQA delivers robust and interpretable quality scores. 
    Extensive experiments on multiple benchmarks demonstrate that our method achieves state-of-the-art performance on tested datasets and aspects, and exhibits strong generalizability in most cross-dataset settings.
    Code is available at \url{https://github.com/strawhatboy/M3-AGIQA}.
\end{abstract}



\received{20 February 2007}
\received[revised]{12 March 2009}
\received[accepted]{5 June 2009}

\maketitle

\section{Introduction}
{I}{n} recent years, flourishing of Artificial Intelligence Generated Content (AIGC) has sparked significant advancements in modalities such as text, image, audio, and even video. 
Among these, AI-Generated Image (AIGI) has garnered considerable interest from both researchers and the public.
Plenty of remarkable text-to-image (T2I) models and online services, such as StableDiffusion\footnote{\url{https://stability.ai/}}, Midjourney\footnote{\url{https://www.midjourney.com/}}, and FLUX\footnote{\url{https://blackforestlabs.ai/}}, offer users an excellent creative experience.
However, users often critique the quality of AIGI results due to visual distortions and misalignment with user intent.
Consequently, methods for assessing the quality of AIGI are becoming increasingly crucial to help improve the generative capabilities of these models.

Unlike Natural Scene Image (NSI) quality assessment, which focuses primarily on perception aspects such as sharpness, color, and brightness, AI-Generated Image Quality Assessment (AGIQA) encompasses additional aspects like correspondence and authenticity. 
Since AIGI is generated on the basis of user text prompts, it may fail to capture key user intentions, resulting in misalignment with the prompt.
Furthermore, authenticity refers to how closely the generated image resembles real-world artworks, as AIGI can sometimes exhibit logical inconsistencies.
While traditional IQA models may effectively evaluate perceptual quality, they are often less capable of adequately assessing aspects such as correspondence and authenticity.

\begin{figure}
    \centering
    \includegraphics[width=0.6\linewidth]{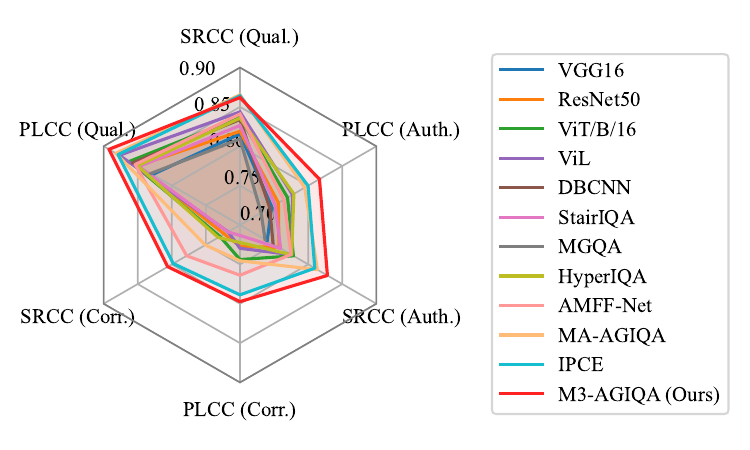}
    \caption{Performance comparison of various image quality assessment methods on the AIGCIQA2023~\cite{wang2023aigciqa2023} dataset across three key aspects: quality (Qual.), correspondence (Corr.), and authenticity (Auth.). The plot reports Spearman’s Rank-Order Correlation Coefficient (SRCC) and Pearson’s Linear Correlation Coefficient (PLCC) metrics, demonstrating that M3-AGIQA achieves higher correlation with human Mean Opinion Score (MOS) than existing baselines, indicating superior alignment with human perceptual judgments.}
    \label{fig:radar}
\end{figure}

Several methods have been proposed specifically for the AGIQA task, including metrics designed to evaluate the authenticity and diversity of generated images~\cite{gulrajani2017improved,heusel2017gans}. 
Nevertheless, these methods tend to compare and evaluate grouped images rather than single instances, which limits their utility for single image assessment.
Beginning with AGIQA-1k~\cite{zhang2023perceptual}, a series of AGIQA databases have been introduced, including AGIQA-3k~\cite{li2023agiqa}, AIGCIQA2023~\cite{wang2023aigciqa2023} and AIGIQA-20k~\cite{li2024aigiqa}, etc.
Concurrently, there has been a surge in research utilizing deep learning methods~\cite{zhou2024adaptive,peng2024aigc,yu2024sf}, which have significantly benefited from pre-trained models such as CLIP~\cite{radford2021learning}. 
These methods enhance the analysis by leveraging the correlations between images and their descriptive texts.
While these models are effective in capturing general text-image alignments, it is difficult for them to detect subtle inconsistencies or mismatches between the generated image content and the detailed nuances of the textual description.
Moreover, as these models are pre-trained on large-scale datasets for broad tasks, they might not fully exploit the textual information pertinent to the specific context of AGIQA without task-specific fine-tuning.
To overcome these limitations, methods that leverage Multimodal Large Language Models (MLLMs)~\cite{wang2024large,wang2024understanding} have been proposed.
These methods aim to fully exploit the synergies of image captioning and textual analysis for AGIQA.
Although these methods benefit from advanced prompt understanding, instruction following, and generation capabilities, they often fall short in fully integrating both image and textual context at a representation level, which can limit their effectiveness in nuanced quality assessment tasks.

In summary, the field of AGIQA continues to face significant challenges: 
(1) How to comprehensively assess AI-generated images from multiple aspects, including perceptual quality, prompt correspondence, and authenticity?
(2) How to develop assessment techniques that more accurately reflect human perception and the nuanced intentions embedded within prompts?
(3) How to fully exploit the multimodal encoding capabilities of MLLMs for more effective and interpretable image quality assessment?

To address these challenges, we propose a novel method M3-AGIQA (\textbf{M}ultimodal, \textbf{M}ulti-Round, \textbf{M}ulti-Aspect \textbf{A}I-\textbf{G}enerated \textbf{I}mage \textbf{Q}uality \textbf{A}ssessment) which leverages MLLMs as both image and text encoders. 
This method incorporates an additional network to align human perception and intentions, aiming to enhance assessment accuracy. 
By harnessing the advanced reasoning and perceptual abilities of MLLMs, our method enables more holistic and human-aligned quality assessment of AI-generated images.
Furthermore, M3-AGIQA introduces a multi-round evaluation strategy, utilizing intermediate image descriptions to better capture key aspects such as quality, correspondence, and authenticity.
The effectiveness of our framework is supported by the integration of knowledge distilled from strong online MLLMs into a local, open-source MLLM, guided by human-labeled data.
The key contributions of this paper are as follows:
\begin{itemize}
    \item We introduce M3-AGIQA, a unified framework that leverages MLLMs and a structured multi-round evaluation process. 
    This enables comprehensive, multi-aspect assessment of AI-generated images, covering not only visual quality but also prompt correspondence and authenticity.
    \item We develop a knowledge distillation pipeline that transfers multi-aspect image captioning capabilities from online MLLMs to a local model through two-round conversational supervision. 
    In addition, we design an xLSTM-based predictor, followed by a regression head, to align model predictions with human perceptual scores by fully exploiting the sequential representations produced by the MLLM.
    \item Extensive experiments across multiple datasets demonstrate that our method achieves superior performance, setting a new state-of-the-art (SOTA) benchmark in AGIQA.
\end{itemize}

In this work, we present related works in Sec.~\ref{sec:related}, followed by the details of our M3-AGIQA method in Sec.~\ref{sec:method}. Sec.~\ref{sec:exp} outlines our experimental design and presents the results. Sec.~\ref{sec:limit},~\ref{sec:ethics} and~\ref{sec:conclusion} discuss the limitations, ethical concerns, future directions and conclusions of our study.

\section{Related Work} \label{sec:related}
\subsection{Blind Image Quality Assessment}
\begin{figure}
    \centering
    \includegraphics[width=0.8\linewidth]{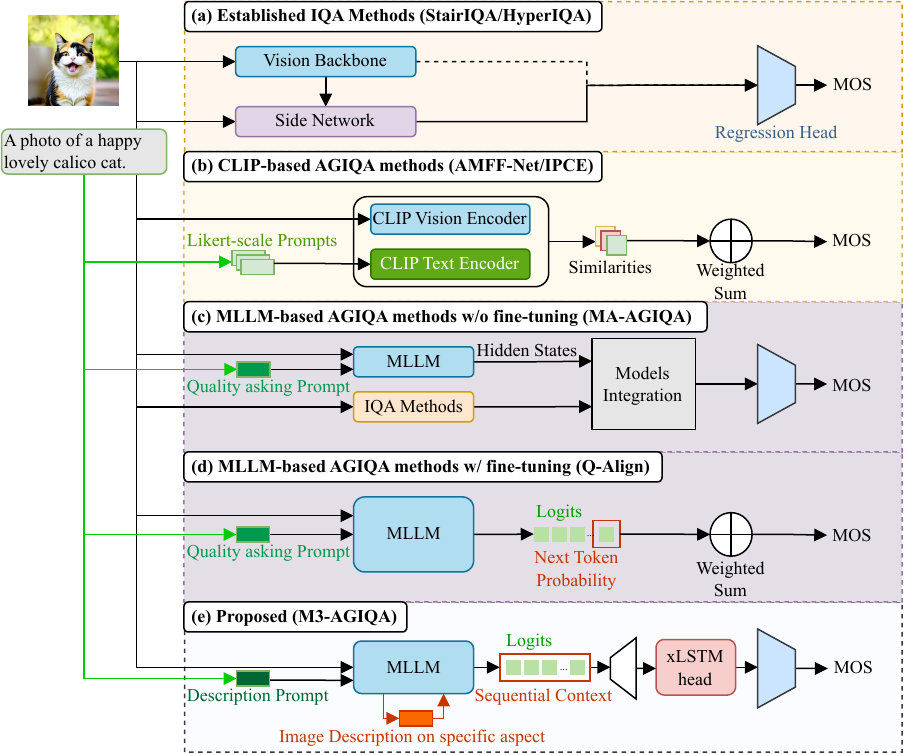}
    \caption{Overview of various image quality assessment (IQA) and AI-generated image quality assessment (AGIQA) methods.
    (a) Established IQA methods that primarily rely on vision backbones and regression heads for quality prediction.
    (b) CLIP-based AGIQA methods incorporate joint vision and text encoders to evaluate image-prompt correspondence and quality via similarity metrics.
    (c) MLLM-based AGIQA methods without fine-tuning utilize pretrained Multimodal Large Language Models (MLLMs) to analyze images and prompts but do not fully exploit MLLM adaptation for the task.
    (d) MLLM-based AGIQA methods with fine-tuning enhance performance by aligning MLLM outputs to human judgments via additional training.
    (e) Our proposed M3-AGIQA framework fine-tunes an open-source MLLM with distillation from online MLLMs, integrates sequential processing of vocabulary logits using an xLSTM head, and employs a regression head to project these rich multimodal representations into continuous MOS. This comprehensive multimodal, multi-round, and multi-aspect approach effectively captures quality, correspondence, and authenticity of AI-generated images.}
    \label{fig:related_work}
\end{figure}
For NSIs, numerous studies over the past few decades have focused on IQA task using various methods. 
Depending on whether a referenced image is used during assessment, these methods can be classified into Full-Reference (FR) and No-Reference (NR) IQA, with NR IQA being more challenging and often referred to as Blind IQA (BIQA).
Traditional BIQA methods, such as NIQE~\cite{mittal2012making}, utilize spatial Natural Scene Statistics (NSS) from a statistical perspective. 
Subsequently, similar works have been developed~\cite{xue2014blind,zhang2014blind,zhang2015feature,xu2016blind} following NIQE.
A significant advancement over these handcrafted feature-based methods is the adoption of deep learning, which has demonstrated superior performance in BIQA tasks~\cite{hou2014blind,kang2014convolutional,zhang2018blind,yang2020ttl,yang2020blind,sun2022blind,zhu2022blind,sun2022graphiqa,tang2023unifying,yang2024joint}.
For example, DBCNN~\cite{zhang2018blind} employs two Convolutional Neural Networks (CNNs) to address both synthetic and authentic distortions. 
HyperIQA~\cite{Su_2020_CVPR} introduces a self-adaptive hyper network that addresses real-world variations through a three-stage process: content understanding, perception rule learning, and quality prediction.
StairIQA~\cite{sun2022blind} integrates information hierarchically from different stages of a ResNet~\cite{he2016deep} backbone with a staircase structure.
As vision-language pre-trained models like CLIP~\cite{radford2021learning}, BLIP~\cite{li2022blip}, and ViT~\cite{dosovitskiy2020vit} gain popularity, new methods~\cite{wang2023exploring,zhang2023blind,shi2023blind,yang2024auxiliary} have emerged that leverage these models to evaluate image quality by assessing text-image similarity.
CLIP-IQA~\cite{wang2023exploring} calculates the similarity between an image and two quality-implied prompts to align with human-labeled scores.
Similarly, LIQE~\cite{zhang2023blind} introduces a multitask learning approach with an expanded set of textual templates and utilizes CLIP to evaluate vision-text similarities. TempQT~\cite{shi2023blind} leverages Transformer~\cite{vaswani2017attention} encoder and decoder to pre-train the error map and then a feature fusion is done by a vision Transformer to predict the quality score.

\subsection{AI-Generated Image Quality Assessment}
With advancements in text-to-image synthesis, assessing the quality of AIGIs has become increasingly important to align with the human visual system (HVS) and the intent implied in the generation prompt.
HPS~\cite{wu2023better} and PickScore~\cite{kirstain2023pick} train a CLIP-based function to predict user preferences by selecting the most preferred one from a group or a pair of AIGIs.
ImageReward~\cite{xu2024imagereward} developed the first text-to-image reward model based on BLIP~\cite{li2022blip}, using a dataset of 137k prompt-image pair rankings sampled from DiffusionDB~\cite{wang2023diffusiondb}, which contains a vast collection of prompt-image pairs without scores. 
Some research pays attention specifically on artistry aspect such as ArtScore~\cite{chen2024learning}, which could be incomplete for AGIQA task.
Then, several datasets using MOS as the evaluation target which presents with human perference well, are introduced to support AGIQA task: AGIQA-1k~\cite{zhang2023perceptual}, AGIQA-3k~\cite{li2023agiqa}, AIGCIQA2023~\cite{wang2023aigciqa2023}, AIGIQA-20k~\cite{li2024aigiqa}, and I2IQA~\cite{yuan2023pku}. 
Based on these datasets, more methods that perform much better have emerged.
AMFF-Net~\cite{zhou2024adaptive} evaluates AIGIs from quality, correspondence and authenticity aspects by scaling images up and down to capture the global and local features, then utilizes CLIP as the text and image encoders.
During the NTIRE2024 Quality Assessment of AI-Generated Content Challenge~\cite{liu2024ntire}, further advancements were made.
Inspired by LIQE~\cite{zhang2023blind}, IPCE~\cite{peng2024aigc} utilizes CLIP to encode quality-aware prompts that include both the quality label and the image prompt, achieving first place in the challenge.
SF-IQA~\cite{yu2024sf}, employing a multilayer feature extractor and fusion module, excels in identifying local and global quality-aware features.
Moreover, MoE-AGIQA~\cite{yang2024moe} sets new benchmarks by integrating visual degradation-aware and semantic-aware networks with a mixture-of-experts module.

While these methods benefit from vision-language models, they typically underutilize the rich textual context including both prompts and intermediate responses, thereby limiting their ability to perform nuanced and human-aligned quality assessment.

\subsection{Multimodal Large Language Models IQA}
Despite great advancements in models like CLIP, which can align images with text, recent methods have started to explore the potential of Multimodal Large Language Models (MLLMs) for the IQA task.
MLLMs inherit strong reasoning and instruction-following capabilities of Large Language Models (LLMs) and can serve as powerful image evaluators when given appropriate prompts.
TIFA~\cite{hu2023tifa} introduces a metric to evaluate the faithfulness of text-to-image generation by using LLMs to generate relevant questions for existing Visual Question Answering (VQA) methods.
LLMScore~\cite{lu2024llmscore} utilizes multi-granularity compositionality capture to evaluate the correspondence between the image and the text prompt by leveraging LLMs as image descriptor and evaluator.
Additionally, Q-Bench~\cite{wu2023q} proposes a systematic benchmark to measure the low-level visual perception and understanding abilities of MLLMs, employing a simple softmax pooling strategy to quantitatively assess the text-image correspondence.
Q-Instruct~\cite{wu2024q} builds upon the same Softmax pooling strategy by constructing 200k instruction-response pairs related to low-level visual attributes.
Q-Align~\cite{wu2024qalign} teaches MLLM to judge based on discrete text levels by converting human rating scores into text, and demonstrates an advantage over score-level variants.
While these methods address general IQA tasks, specialized approaches for AGIQA are also emerging.
MA-AGIQA~\cite{wang2024large} integrates semantically informed guidance with quality-aware features through a Mixture of Experts (MoEs) structure, leveraging both MLLM and traditional DNN approaches for superior performance.
MINT-IQA~\cite{wang2024understanding} extends AIGCIQA2023~\cite{wang2023aigciqa2023} database to AIGCIQA2023+ by adding image quality descriptions and applying instruction tuning to MLLMs, resulting in considerable improvements in assessing human visual preferences from multiple perspectives.

Although these methods utilize MLLM abilities like prompt understanding, instruction following and text generation, they often overlook the potential encoding capability of MLLMs.

\subsection{Long Short-Term Memory Network} \label{subsec:lstm}
Recurrent Neural Networks (RNNs) are foundational for sequence modeling but suffer from issues such as vanishing and exploding gradients, which hinder their ability to capture long-term dependencies.
Long Short-Term Memory networks (LSTMs)~\cite{hochreiter1997long} were developed to address these problems by introducing gated memory cells, enabling more effective learning over extended sequences. 
However, standard LSTMs can still face challenges with very long sequences and lack efficient parallelizability.
To further enhance sequential modeling, the extended LSTM (xLSTM)~\cite{poppel2024xlstm} architecture was proposed. xLSTM introduces exponential gating and advanced memory structures, making it competitive with state-of-the-art models such as Transformers and State Space Models (SSMs).
The xLSTM transformation applied to sequential features can be formulated as:
\begin{equation}
    \boldsymbol{H}_l = \textrm{xLSTM}_l\left(\boldsymbol{H}_{l-1}\right),
\end{equation}
where $\boldsymbol{H}_{l-1}$ is the input feature sequence to the $l$-th xLSTM block, and $\boldsymbol{H}_l$ is the output feature sequence. Each layer $l$ corresponds to either an mLSTM or sLSTM block, as defined in~\cite{poppel2024xlstm}. 
This modular stacking allows xLSTM to effectively capture both short-term and long-term dependencies with linear computational and memory complexity.
Building upon xLSTM, VisionLSTM (ViL)~\cite{alkin2024vision} adapts these blocks for vision tasks by stacking alternating mLSTM and sLSTM layers to efficiently process sequences of image patches, demonstrating robust performance as a general-purpose vision backbone.
In this work, we employ a four-block xLSTM head (mLSTM, sLSTM, mLSTM, sLSTM) as our sequential feature extractor. 
This design enables effective modeling of the complex dependencies present in the sequential vocabulary logits produced by our fine-tuned MLLM, leading to improved quality assessment accuracy.

\section{Method} \label{sec:method}
\begin{figure*}
    \centering
    \includegraphics[width=1\linewidth]{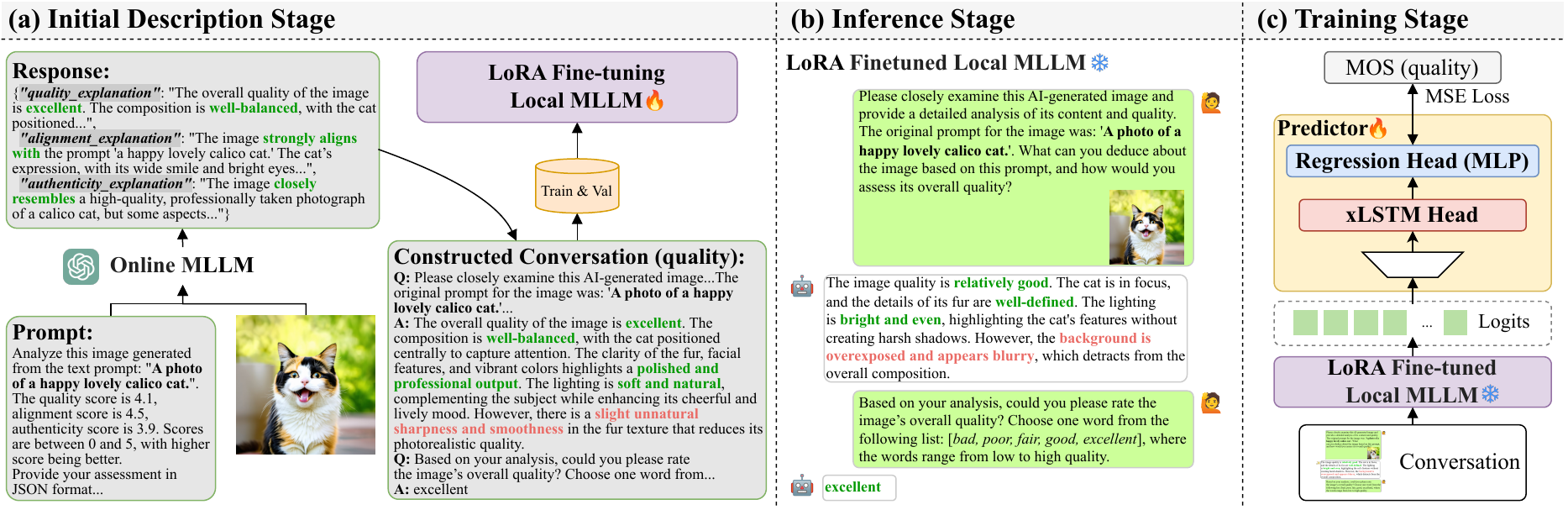}
    \caption{Overview of the M3-AGIQA framework. This diagram illustrates the three-stage process using the quality aspect of AIGIs as an example. \textbf{(a) Initial Description Stage:} By using an online MLLM service via API, we obtain descriptions related to quality, correspondence, and authenticity for each image. 
    These descriptions are then used to fine-tune an open-source MLLM, enhancing its understanding of these aspects.
    \textbf{(b) Inference Stage:} Distilled captioning capabilities of the fine-tuned MLLM were leveraged coupled with the textual prompt to generate the required response. This step utilizes the refined ability of the MLLM to interpret and articulate the nuances of the input prompt and image.
    \textbf{(c) Training Stage:} We freeze all the parameters within the fine-tuned MLLM to ensure stability and reproducibility. An xLSTM head along with a regression head is then trained to predict the MOS, effectively translating enhanced perceptual and textual understanding of the MLLM into quantifiable assessments.}
    \label{fig:overall}
\end{figure*}

Our work introduces a novel framework to address the AGIQA task, which seeks to establish a function $f$ that accepts an AIGI $i$ and its corresponding text prompt $p$ as inputs, and outputs a predicted MOS:
\begin{equation}
\hat{y} = f\left(i, p\right).
\end{equation}
Since some datasets provide MOS for different aspects, we design separate predictive functions for each dimension of interest. 
Specifically, $f_q$ is used for quality, $f_c$ for correspondence, and $f_a$ for authenticity. 
This aspect-specific modeling enables a more nuanced and human-aligned evaluation of AI-generated images.
As illustrated in Fig.~\ref{fig:overall}, M3-AGIQA follows a three-stage process: 
(1) an initial description stage that generates aspect-specific captions using an online MLLM; 
(2) an inference stage that employs multi-round conversations with a fine-tuned local MLLM to facilitate deeper understanding; 
and (3) a training stage that learns to predict MOS from the MLLM outputs.

\subsection{Initial Description Stage} \label{subsec:distil}
The initial description stage generates high-quality descriptions for quality, correspondence, and authenticity aspects of AIGIs, forming the foundation for downstream training.
Distillation from online MLLM, which excels in image understanding~\cite{jin2024efficient}, transfers its capabilities to an open-source MLLM, reducing cost and latency while maintaining performance.

In the first stage of our method, we use the online MLLM to generate descriptions of AIGIs from three aspects: quality, correspondence, and authenticity.
To achieve this, we manually create a prompt that is then refined by \textit{GPT-4}, and the prompt embeds both the MOS and the specific text prompt of the AIGI (See Appendix~\ref{app:prompt_getid}).
The dual embedding guides the MLLM in producing descriptions that not only resonate with human perceptions but also exploit the object detection capabilities of the model to enhance text-image alignment judgments.
Then, we call the API of the online MLLM $\mathcal{G}$ to generate descriptions of the AIGIs from the datasets.
\begin{equation}
    \boldsymbol{D^i} = \mathcal{G}_\textrm{gen}\left(i, \textrm{T}_\textrm{desc}\left(p, y\right)\right),
\end{equation}
\begin{equation}
    \boldsymbol{D^i} = \left\{D^i_q, D^i_c, D^i_a\right\},
\end{equation}
where $\boldsymbol{D^i}$ is the description which consists of descriptions on three aspects $D^i_q, D^i_c, D^i_a$ for AIGI $i$; $\mathcal{G}_\textrm{gen}$ denotes the generate function of the online MLLM API; and $\textrm{T}_\textrm{desc}$ is the prompt template which accepts the text prompt $p$ and MOS $y$ of the AIGI as inputs.

After generating descriptions for the AIGIs, we initiate a two-round conversational interaction between a \textit{user} and an \textit{assistant} with a new prompt template (See Appendix~\ref{app:prompt_multiround}), as depicted in Fig.~\ref{fig:overall}~(a).
This template incorporates Likert-scale of five-level descriptors [\textit{bad, poor, fair, good, excellent}] as part of the prompt construction, following methodologies similar to those of previous works~\cite{ghadiyaram2015massive, zhang2023blind, wu2024qalign, zhuadaptive}.

\begin{figure}
    \centering
    \includegraphics[width=1.0\linewidth]{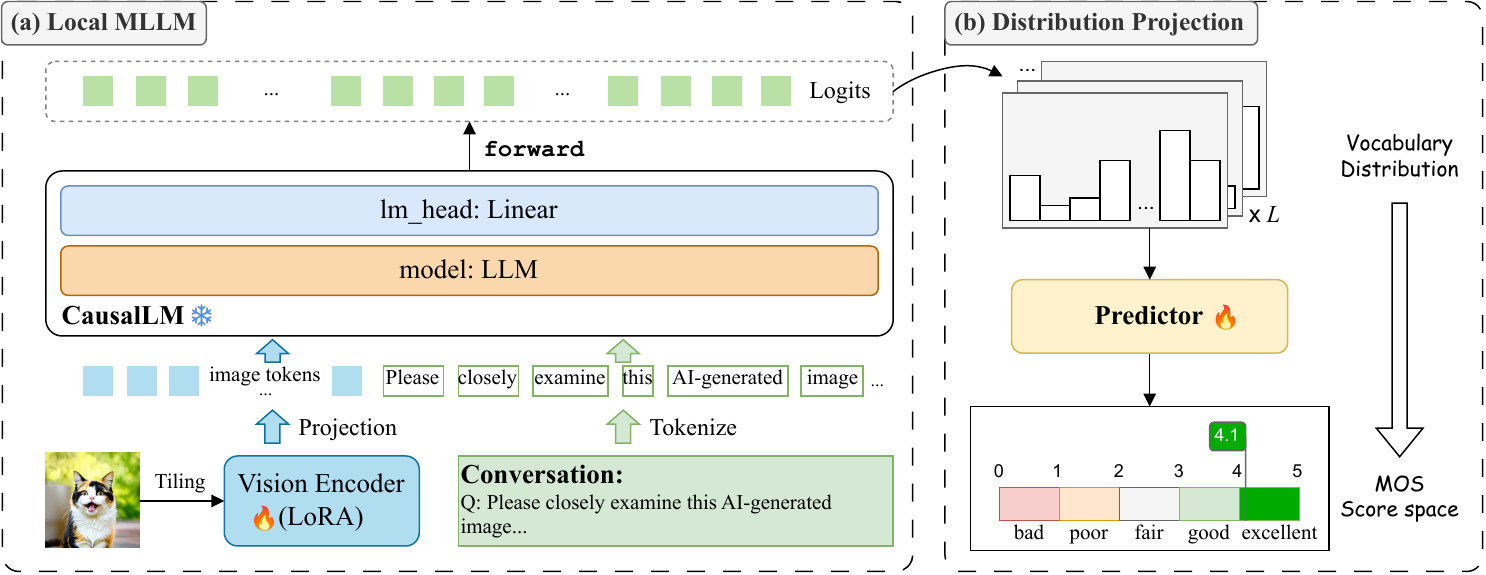}
    \caption{(a) The local MLLM is fine-tuned via LoRA on the vision encoder using input images and conversations. The MLLM outputs a sequence of vocabulary logits after processing visual and textual tokens.
    (b) These logits serve as rich sequential features, which are then processed by an xLSTM head and a regression head to project into continuous MOS aligned with human perceptual judgments. }
    \label{fig:localmllm}
\end{figure}

Then, the conversations and their corresponding AIGIs are divided into training and validation sets.
These sets are then fed into a local open-source MLLM for training with Low-Rank Adaptation (LoRA)~\cite{hu2022lora} fine-tuning technique on the vision encoder only inside the MLLM as illustrated in Fig.~\ref{fig:localmllm} (a).
LoRA introduces trainable low-rank matrices into the linear layers of the vision encoder, allowing efficient adaptation while freezing most of the original pre-trained weights. Formally, for a linear layer with weight matrix $\boldsymbol{W} \in \mathbb{R}^{d \times k}$, LoRA reparameterizes it as:
\begin{equation}
    \boldsymbol{W'}=\boldsymbol{W} + \boldsymbol{A}\boldsymbol{B},
\end{equation}
where $\boldsymbol{A} \in \mathbb{R}^{d \times r}$ and $\boldsymbol{B} \in \mathbb{R}^{r \times k}$ are low-rank matrices with rank $r \ll d,k$. During fine-tuning, only $\boldsymbol{A}$ and $\boldsymbol{B}$ are updated, whereas the original weights $\boldsymbol{W}$ remain frozen. 
This parameter-efficient approach allows us to adapt the MLLM to the AGIQA task with reduced computational cost and memory footprint.

Initially, the AIGI is resized and segmented into patches before being sent to the vision encoder, after which the encoded data for each patch are projected into visual tokens.
These visual tokens are combined with text tokens and input into the LLM.
Inside the LLM, all the tokens are processed, and the linear layer at the end outputs the logits of the sequence, which can be subsequently decoded into text.
Thus, we can obtain the fine-tuned local MLLM $\mathcal{F}$, which accepts AIGI $i$ and text prompt $p$ as inputs.

\subsection{Inference Stage} \label{subsec:inference}
In the inference stage, we harness the zero-shot Chain of Thought (CoT) capabilities of LLMs~\cite{kojima2022large} to generate intermediate reasoning processes that refine the final conclusions.
As illustrated in Fig.~\ref{fig:overall}~(b), this stage enhances the accuracy of assessments by leveraging the generated descriptions to judge the quality of AIGIs, while the two-round conversation process augments the data for subsequent training.
The fine-tuned local open-source MLLM in Sec.~\ref{subsec:distil} produces meaningful logits during this process, representing the distribution of the AIGI and text across vocabulary probabilities, which are crucial for accurately predicting the MOS.
Skipping this step results in inadequate understanding in image assessment and disrupts the connection between distilled knowledge and the final score prediction.
The process of generating description $D^i_\textrm{ft}$ can be simplified as follows:
\begin{equation}
    D^i_\textrm{ft} = \mathcal{F}_\textrm{gen}\left(i, p\right),
\end{equation}
where $\mathcal{F}_\textrm{gen}$ is the generate function of the fine-tuned local MLLM $\mathcal{F}$.

\subsection{Training Stage}
To align the output of the MLLM with the MOS, which reflects human perceptual assessment, we introduce a predictor network that operates on the sequential representations produced by the MLLM. 
While LLM2Vec~\cite{behnamghaderllm2vec} demonstrates the effectiveness of using a large language model as a text encoder, our method extends this idea by leveraging an MLLM as a unified encoder for both text and image modalities.
As such, we utilize the generated conversation $C^i$, which includes the generated description $D^i_\textrm{ft}$ from the previous inference stage (see Sec.~\ref{subsec:inference}) along with the corresponding AIGI as inputs to encode them as follows:
\begin{equation}
    \boldsymbol{E^i} = \mathcal{F}_\textrm{forw}\left(i, C^i\right),
\end{equation}
where $\boldsymbol{E^i} \in \mathbb{R}^{L \times d_\textrm{vocab}} $ represents the output logits by calling the forward function $\mathcal{F}_\textrm{forw}$ of the fine-tuned local MLLM $\mathcal{F}$, with $L$ denoting the sequence length of the input conversation $C^i$ and $d_\textrm{vocab}$ representing the vocabulary size of MLLM tokenizer.
The last element $\boldsymbol{E^i[-1]}$ of the output logits, typically represents the probability of the next token, which is possibly valuable for predicting the target.
However, as Wang et al.~\cite{wang2024my} mentioned, the next first-token probability might lead to a mismatch with the expected answer. This observation is also supported by our experiments (see Sec.~\ref{sec:variant}). 
Instead, we treat the entire logits sequence $\boldsymbol{E^i}$ as a high-dimensional, sequential feature representation that preserves nuanced relationships across tokens and modalities.
To effectively extract and summarize these features, we employ an xLSTM head (described in Sec.~\ref{subsec:lstm}) which is well-suited for modeling long-range dependencies in sequential data. The process is as follows:
\begin{equation}
    \boldsymbol{E^i_\textrm{proj}} = \mathbf{W}_\textrm{proj} \boldsymbol{E^i},
\end{equation}
\begin{equation}
    \boldsymbol{E^i_\textrm{out}} = \textrm{xLSTMHead}\left(\boldsymbol{E^i_\textrm{proj}}\right),
\end{equation}
\begin{equation}
    \boldsymbol{z^i} = \textrm{MeanPooling}\left(\boldsymbol{E^i_\textrm{out}}\right),
\end{equation}
where $\mathbf{W}_\textrm{proj} \in \mathbb{R}^{d_{\textrm{vocab}} \times d_\textrm{h}}$ represents the learnable linear layer that projects from the vocabulary space to the xLSTM hidden space of dimension $d_\textrm{h}$; $\boldsymbol{E^i_\textrm{proj}}, \boldsymbol{E^i_\textrm{out}} \in \mathbb{R}^{L \times d_\textrm{h}}$ are input and output of the xLSTM head. 

The result embedding $\boldsymbol{z^i} \in \mathbb{R}^{d_\textrm{h}}$ is fed into a regression head implemented as a Multilayer Perceptron (MLP) with two fully connected layers. This head maps the sequential features $\boldsymbol{z^i}$ into a continuous scale $\hat{y}^i$ corresponding to the MOS:
\begin{equation}
    \hat{y}^i = \textrm{RegressionHead}\left(\boldsymbol{z^i}\right).
\end{equation}
The overall mapping from sequential logits to MOS is illustrated in Fig.~\ref{fig:localmllm} (b).
By processing the sequential, multimodal output distribution of the MLLM through the xLSTM head and regression head, our method produces a precise, quantitative quality assessment that is closely aligned with human perception.

To optimize the model, Mean Square Error (MSE) loss is employed for backpropagation through the trainable parameters in the predictor:
\begin{equation}
    L = \frac{1}{n} \sum_{i=1}^{n}{\left(\hat{y}^i - y^i\right)^2},
\end{equation}
where $y^i$ is the ground truth MOS.

\section{Experiments} \label{sec:exp}
This section presents extensive experiments conducted to evaluate our proposed method M3-AGIQA in comparison with other SOTA models.
Our experimental design is structured to address the following key research questions:
\textbf{RQ1:} How does M3-AGIQA compare in performance with current SOTA methods?
\textbf{RQ2:} What is the contribution of each component within M3-AGIQA to its overall performance?
\textbf{RQ3:} Does the distillation process enhance the overall performance of the model?
\textbf{RQ4:} How do component variants influence performance?
\textbf{RQ5:} How does the fine-tuned model perform across different datasets?
\subsection{Setup} \label{subsec:setup}
\subsubsection{Datasets}
As summarized in Table~\ref{tab:dataset}, we utilize three public AGIQA datasets for our experiments, including AGIQA-3k~\cite{li2023agiqa}, AIGCIQA2023~\cite{wang2023aigciqa2023}, and AIGIQA-20k~\cite{li2024aigiqa}.
Each dataset is widely recognized in the field and provides MOS that assess quality, correspondence, and authenticity aspects either fully or partially.
\textbf{AGIQA-3k}~\cite{li2023agiqa} includes $2982$ images generated by $6$ different models which are GAN, auto-regression, and diffusion-based models. It annotates MOS for image perception quality and correspondence with the prompt.
\textbf{AIGCIQA2023}~\cite{wang2023aigciqa2023} is composed of $2400$ images from $6$ cutting-edge models with MOS for three aspects: quality, correspondence, and authenticity. Each prompt generates $4$ images for one model.
\textbf{AIGIQA-20k}~\cite{li2024aigiqa} from the NTIRE 2024 Quality Assessment Challenge has $20,000$ images generated by $15$ popular models, along with MOS collected from 21 subjects.

\begin{table}
    \caption{Statistics of the datasets.\label{tab:dataset}}
    \centering
    \begin{tabular}{cccc}
        \toprule
        Statistics & AGIQA-3k & AIGCIQA2023 & AIGIQA-20k \\
        \midrule
        No. of images & 2,982 & 2,400 & 20,000 \\
        No. of T2I models & 6 & 6 & 15 \\
        quality MOS & \ding{51} & \ding{51} & \ding{51} \\
        correspondence MOS & \ding{51} & \ding{51} & \ding{55} \\
        authenticity MOS & \ding{55} & \ding{51} & \ding{55} \\ 
        \bottomrule
    \end{tabular}
\end{table}

\subsubsection{Baselines}
To demonstrate the effectiveness of our proposed method M3-AGIQA, we select several baselines for comparative analysis: 
(1) \textbf{Simple vision encoders with regression head}: ResNet50~\cite{he2016deep}, ViT/B/16~\cite{dosovitskiy2020vit}, and ViL~\cite{alkin2024vision}. We integrate a two-layer MLP as the regression head to directly predict the MOS; 
(2) \textbf{Established IQA methods}: DBCNN~\cite{zhang2018blind}, HyperIQA~\cite{Su_2020_CVPR}, StairIQA~\cite{sun2022blind}, and MGQA~\cite{wang2021multi}; 
(3) \textbf{AGIQA methods}: AMFF-Net~\cite{zhou2024adaptive}, MA-AGIQA~\cite{wang2024large}, and IPCE~\cite{peng2024aigc}. 
\subsubsection{Implementation Details}
The MLLM used in the experiments is \textit{openbmb/MiniCPM-Llama3-V-2\_5}~\cite{yao2024minicpmv}, a lightweight \textit{GPT-4V} level MLLM. 
To generate image descriptions for distillation, we utilize the free online MLLM Google \textit{gemini-1.5-pro} API as the teacher model.
The distillation was conducted via an official fine-tuning script\footnote{\url{https://github.com/OpenBMB/MiniCPM-V}}. 
Our xLSTM head comprises four layers: three mLSTM layers (at positions $1$, $3$, and $4$) and one sLSTM layer (at position $2$). The regression head consists of two fully connected layers.
The total number of parameters in the predictor (including the projection layer, the xLSTM head and the regression head) is approximately $70M$, compared to the approximately $8$B parameters of the local MLLM.
Our experiments employed \textit{PyTorch} and \textit{TorchLightning 2.3.0}\footnote{\url{https://lightning.ai/docs/pytorch/2.3.0/}} to implement the training process.
The datasets were partitioned based on practices established in previous studies: 4:1:0 for AGIQA-3k~\cite{li2023agiqa}, 3:1:0 for AIGCIQA2023~\cite{wang2023aigciqa2023}, and 7:2:1 for AGIQA-20k, in terms of training, test, and validation sets respectively.
During fine-tuning, the learning rate was set to $2e$-$6$, and the batch size was fixed at $2$.
The vision encoder was fine-tuned using the LoRA technique while the LLM parameters remained frozen.
In addition, deepspeed ZeRO-3 offload was employed to minimize GPU VRAM usage.
The fine-tuning process ranged from $2,000$ to $4,000$ steps for AGIQA-3k~\cite{li2023agiqa} and AIGCIQA2023~\cite{wang2023aigciqa2023}, and around $20,000$ steps for AIGIQA-20k~\cite{li2024aigiqa}.
In the training stage, AdamW was utilized as the optimizer.
The hidden dimension size $d$ was set to $512$ and the vocabulary size of the MLLM tokenizer $d_{vocab}$ was $128256$.
All experiments were conducted using an NVIDIA A100-PCIE-40GB GPU and an Intel Xeon Gold 6248R CPU.
\subsubsection{Metrics}
To evaluate the performance of our method, we utilized two widely-used metrics in IQA tasks: Spearman's Rank-Order Correlation Coefficient (SRCC) and Pearson's Linear Correlation Coefficient (PLCC).
SRCC measures ability of the model to preserve the rank order of the predictions relative to the ground truth, indicating its effectiveness in ranking images based on quality.
PLCC evaluates the linear correlation between the predicted and actual scores, representing how the model fits the data.
Both metrics range from $-1$ to $1$, with higher values indicating better performance.

\subsection{Comparison with SOTAs - RQ1}
\begin{table*}
    \centering
    \caption{Comparison results on AGIQA-3k~\cite{li2023agiqa}, AIGCIQA2023~\cite{wang2023aigciqa2023}, and AIGIQA-20k~\cite{li2024aigiqa} among different methods, results of methods with asterisk symbol ``$^\ast$'' are directly retrieved from corresponding papers. \textbf{Bold} and \underline{underlined} values indicate the best and second-best results, respectively.}
    \label{tab:comparison}
    \resizebox{\textwidth}{!}{
    \begin{tabular}{c|cc cc cc cc cc cc|cc}
        \toprule
        \multirow{3}{*}{Methods} & \multicolumn{4}{c|}{AGIQA-3k} & \multicolumn{6}{c|}{AIGCIQA2023} & \multicolumn{2}{c|}{AIGIQA-20k} & \multicolumn{2}{c}{\multirow{2}{*}{Average}} \\
        \cline{2-5} \cline{6-9} \cline{10-13}
        & \multicolumn{2}{c|}{Qual.} & \multicolumn{2}{c|}{Corr.} & \multicolumn{2}{c|}{Qual.} & \multicolumn{2}{c|}{Corr.} & \multicolumn{2}{c|}{Auth.} & \multicolumn{2}{c|}{Qual.} \\
        \cline{2-15}
        & SRCC & PLCC & SRCC & PLCC & SRCC & PLCC & SRCC & PLCC & SRCC & PLCC & SRCC & PLCC & SRCC & PLCC \\
        \midrule
        VGG16~\cite{Simonyan15} & 0.8167 & 0.8752 & 0.6867 & 0.8108 & 0.8157 & 0.8282 & 0.6839 & 0.6853 & 0.7399 & 0.7465 & 0.8133 & 0.8660 & 0.7594 & 0.8020 \\
        ResNet50~\cite{he2016deep} & 0.8552 & 0.9072 & 0.7493 & 0.8564 & 0.8190 & 0.8503 & 0.7230 & 0.7270 & 0.7571 & 0.7563 & 0.8036 & 0.8661 & 0.7845 & 0.8272 \\

        ViT/B/16~\cite{dosovitskiy2020vit} & 0.8659 & 0.9115 & 0.7478 & 0.8449 & 0.8370 & 0.8618 & 0.7293 & 0.7439 & 0.7783 & 0.7697 & 0.8407 & 0.8904 & 0.7998 & 0.8370  \\
        ViL~\cite{alkin2024vision} & 0.8750 & 0.9145 & 0.7570 & 0.8561 & 0.8436 & 0.8770 & 0.7174 & 0.7296 & 0.7753 & 0.7770 & 0.8459 & 0.8852 & 0.8024 & 0.8399 \\

        \midrule
        DBCNN$^\ast$~\cite{zhang2018blind} & 0.8154 & 0.8747 & 0.6329 & 0.7823 & 0.8339 & 0.8577 & 0.6837 & 0.6787 & 0.7485 & 0.7436 & 0.7941 & 0.8542 & 0.7514 & 0.7985 \\
        StairIQA~\cite{sun2022blind} & 0.8439 & 0.8989 & 0.7345 & 0.8474 & 0.8264 & 0.8483 & 0.7176 & 0.7133 & 0.7596 & 0.7514 & 0.8126 & 0.8746 & 0.7824 & 0.8223\\

        MGQA~\cite{wang2021multi} & 0.8283 & 0.8944  & 0.7244 & 0.8430 & 0.8093 & 0.8308 & 0.6892 & 0.6745 & 0.7367 & 0.7310 & 0.8107 & 0.8534 & 0.7664 & 0.8045 \\
        HyperIQA~\cite{Su_2020_CVPR} & 0.8526 & 0.8975 & 0.7437 & 0.8471 & 0.8357 & 0.8504 & 0.7318 & 0.7222 & 0.7758 & 0.7790 & 0.8171 & 0.8584 & 0.7928 & 0.8258\\
        \midrule
        AMFF-Net$^\ast$~\cite{zhou2024adaptive} & 0.8565 & 0.9050 & 0.7513 & 0.8476 & 0.8409 & 0.8537 & 0.7782 & 0.7638 & 0.7749 & 0.7643 & - & - & 0.8004 & 0.8269 \\
        MA-AGIQA~\cite{wang2024large} & 0.8709 & 0.9136 & \underline{0.7721} & \underline{0.8785} & \textbf{0.8657} & \underline{0.8821} & 0.7509 & 0.7455 & \underline{0.8134} & 0.7949 & 0.8619 & 0.9006 & 0.8225 & 0.8525 \\

        IPCE~\cite{peng2024aigc} & \underline{0.8841} & \underline{0.9246} & 0.7697 & 0.8725 & \underline{0.8640} & 0.8788 & \underline{0.7979} & \underline{0.7887} & 0.8097 & \underline{0.7998} & \textbf{0.9076} & \underline{0.9274} & \underline{0.8388} & \underline{0.8653} \\
        \midrule
        M3-AGIQA (Ours) & \textbf{0.9045} & \textbf{0.9317} & \textbf{0.8523} & \textbf{0.9142} & 0.8618 & \textbf{0.8922} & 
        \textbf{0.8060} & \textbf{0.7973} & \textbf{0.8282} & \textbf{0.8165} & \underline{0.8988} & \textbf{0.9292} & \textbf{0.8586} & \textbf{0.8802} \\
        \bottomrule
    \end{tabular}
    }
\end{table*}

\begin{figure*}
    \centering
    \begin{subfigure}{.48\linewidth}
        \includegraphics[width=\textwidth]{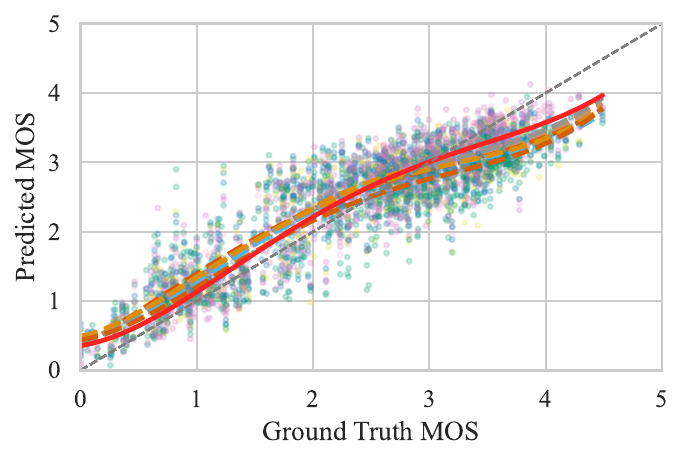}
        \caption{Predicted vs. Ground Truth MOS}
    \end{subfigure}
    \begin{subfigure}{.48\linewidth}
        \includegraphics[width=\textwidth]{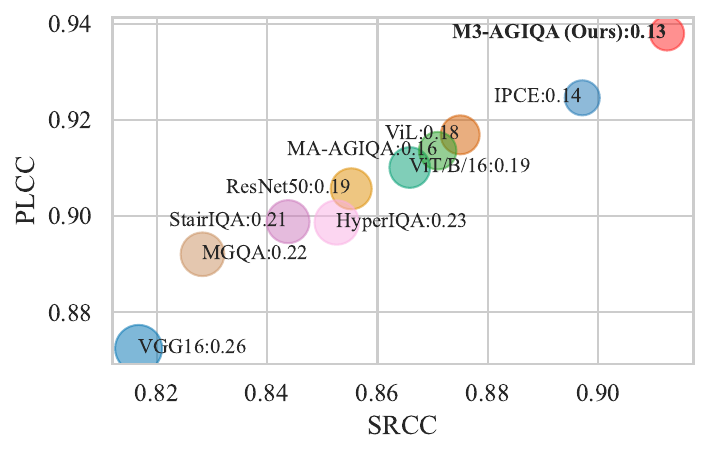}
        \caption{Correlation Coefficients}
    \end{subfigure}

    \caption{Comparison of model performance on the quality aspect of the AGIQA-3k~\cite{li2023agiqa} dataset. (a) Scatter plot of predicted MOS versus ground truth MOS for all reproducible methods. Fitted curves for each method are obtained using fourth-order polynomial regression; the red solid line represents our method and is closer to the ideal diagonal, indicating superior fit. (b) SRCC vs. PLCC plot for all reproducible methods, where the size of each dot corresponds to the MSE during prediction. Smaller dots and positions closer to the top-right corner denote better overall performance.}
    
    \label{fig:scatter}
\end{figure*}

Experiments on the datasets have shown our proposed method M3-AGIQA significantly outperforms the counterparts. As depicted in Table~\ref{tab:comparison},
simple vision encoders perform well on AGIQA task, ViL~\cite{alkin2024vision} which utilizes the latest xLSTM~\cite{poppel2024xlstm} architecture shows strong advancement beyond the traditional ResNet50~\cite{he2016deep}. 
With respect to the traditional BIQA methods, VGG~\cite{Simonyan15} based DBCNN~\cite{zhang2018blind} is simple and fast but not competitive due to a lack of quality related feature extraction ability;
ResNet50~\cite{he2016deep} based StairIQA~\cite{sun2022blind}, MGQA~\cite{wang2021multi}, and HyperIQA~\cite{Su_2020_CVPR} do not improve much or are even inferior to their backbone, this could occur when they are designed specifically from the perspective of the quality of NSIs but not AIGIs, which would ignore the impact from correspondence and authenticity aspects.
AGIQA methods perform significantly better than the simple vision encoders and BIQA methods do, especially in terms of correspondence and authenticity apsects.
As an example on AIGCIQA2023~\cite{wang2023aigciqa2023} dataset, IPCE~\cite{peng2024aigc} which combines quality aware text with the AIGI as input of CLIP~\cite{radford2021learning} model, achieves overall superior results to those of non-AGIQA methods, not only because of the extraordinary ability of CLIP model to align with text and image, but also because of the meaningful text adoption on the three aspects.

Compared with the baselines, M3-AGIQA consistently demonstrates strong or leading performance across major AGIQA datasets and different aspects. 
Notably, M3-AGIQA achieves the highest PLCC scores in most cases, indicating better alignment with human-annotated MOS and a superior fit to ground truth.
While certain baselines such as IPCE, show advantages in terms of specific ranking metrics or scenarios, our method maintains robust performance across both ranking and regression tasks.

Importantly, these results suggest that while certain baselines are effective in specialized settings, M3-AGIQA offers greater generalizability and interpretability because of its multi-aspect, multi-round conversational framework and knowledge distillation strategy. 
This makes M3-AGIQA a more comprehensive and reliable solution for the assessment of AI-generated image quality across diverse conditions.

\subsection{Ablation Study - RQ2, RQ3}
To determine the contributions of each component to the overall performance of our method, experiments were conducted at every stage of the process.

\subsubsection{Image Descriptions}\label{sec:img_desc}
To demonstrate the importance of inferred image descriptions in terms of quality, correspondence, and authenticity, and their roles as contextual inputs for enhancing deep understanding of the AIGI, we assessed their impact specifically on the quality aspect of dataset AGIQA-3k~\cite{li2023agiqa}, with results detailed in Table~\ref{tab:img_desc}.
The column ``Present Descriptions'' denotes the various combinations of generated image descriptions used throughout the experimental stages.
The first line, which lacks any descriptions, depicts a scenario where the model directly predicts the outcome without any contextual input. 
This configuration suggests we only fine-tuned the local MLLM with a single-round conversation, excluding any detailed responses from the MLLM, and can be interpreted as \textit{w/o CoT}. 
In other setups, the MLLM was prompted to describe the image across different aspects, thereby enriching the context for the prediction task.

\begin{table}[h]
    \centering
    \caption{Comparison results on AGIQA-3k~\cite{li2023agiqa} quality aspect with different combinations of image description aspects.}
    \label{tab:img_desc}
    \begin{tabular}{ccc| cc}
        \toprule
        \multicolumn{3}{c|}{Present Descriptions} & 
        \multicolumn{2}{c}{Qual.} \\
        \hline
        Qual. & Corr. & Auth. & SRCC & PLCC  \\
        \midrule
        \ding{55} & \ding{55} & \ding{55} & 0.8816 & 0.9193 \\
        \ding{51} & \ding{55} & \ding{55} & 0.8989 & 0.9342 \\
        \ding{51} & \ding{51} & \ding{55} & 0.9045 & 0.9317 \\
        \ding{51} & \ding{51} & \ding{51} & 0.8999 & 0.9321 \\
        \bottomrule
    \end{tabular}
\end{table}

The results reveal that the model trained without any intermediate contextual descriptions has the weakest performance, suggesting that leveraging the innate capabilities of the original MLLM is insufficient.
In contrast, adding descriptions that cover both quality and correspondence aspects leads to the best performance, indicating the correspondence aspect likely contributes to human-like perspective enhancements that positively influence quality judgments.
However, including additional aspects such as authenticity slightly degrades performance. 
This might be because authenticity has a less direct relationship with image quality aspect, and its inclusion introduces noise that could detract from the effectiveness of the model in assessing quality.

\subsubsection{Distillation \& Fine-tuning}
Additional experiments on AGIQA-3k~\cite{li2023agiqa} quality aspect in Table~\ref{tab:ab2} were conducted by including the inference stage but maintained the absence of the initial description stage (w/o ft, w/ ID) in contrast to our final settings (w/ ft, w/ ID). The noticable performance drop highlights the necessity of distilling descriptive knowledge from online MLLM into local MLLM through fine-tuning.

\begin{table}
    \centering
    \caption{Ablation studies on AGIQA-3k~\cite{li2023agiqa} quality aspect.}
    \label{tab:ab2}
    \begin{tabular}{c| cc}
        \toprule
        Model settings & SRCC & PLCC  \\
        \midrule
        w/o ft, w/o ID & 0.8886 & 0.9208 \\
        w/o ft, w/ ID & 0.8800 & 0.9030 \\
        w/o ft, w/ FullConv & 0.8720 & 0.8939 \\
        w/ ft, w/o ID & 0.8921 & 0.9249 \\
        w/ ft, w/ FullConv & 0.8968 & 0.9278 \\
        \midrule
        w/o xLSTM & 0.8959 & 0.9274 \\
        w/o lm\_header & 0.8949 & 0.9325 \\
        \midrule
        
        M3-AGIQA (Ours) & 0.9045 & 0.9317 \\
        \bottomrule
    \end{tabular}
\end{table}

After inferencing the result, we approached the AGIQA task as a classification problem, assigning labels [\textit{bad, poor, fair, good, excellent}] within dataset AGIQA-3k~\cite{li2023agiqa}.
Classification labels are segmented according to ranges of MOS, for example scores from [0-1], [1-2], and [2-3] correspond to bad, poor, and fair, respectively.
Thus, an image with a target label score 1.99 but predicted as 2.01 would be classified as fair instead of poor.
Given these nuances, adopting a less stringent classification metric termed ``Rough Accuracy'' is beneficial:
\begin{equation}
    \textrm{Rough Acc.} = \frac{1}{N}\sum_{i=1}^{N}{\mathbbm{1}\left(\left|\hat{y^i}-y^i\right| \leq 1 \right)},
\end{equation}
where $\mathbbm{1}(\cdot)$ is the indicator function, $\hat{y_i}$ and $y_i$ are the predicted and ground truth quality labels, each taking a value in $\{0,1,2,3,4\}$ representing a quality category. 
$N$ is the total number of samples evaluated. 
As detailed in Table~\ref{tab:ab3cls}, the overall performance is quite favorable, demonstrating the effectiveness of fine-tuning.

\begin{table}[h]
    \centering
    \caption{Classification task on AGIQA-3k~\cite{li2023agiqa} quality aspect.}
    \label{tab:ab3cls}
    \begin{tabular}{c| cccc}
        \toprule
        Model settings & Rough Acc. & Acc. & Precision & F1  \\
        \midrule
        w/o ft, w/ ID & 0.5427 & 0.1005 & 0.2615 & 0.1018 \\
        w/ ft, w/ ID & 0.9916 & 0.6616 & 0.6462 & 0.6534 \\
        \bottomrule
    \end{tabular}
\end{table}

\subsubsection{MLLM as an Encoder}\label{sec:mllm_encoder}
The LLM inside the MLLM utilizes a linear head to project hidden representations to the vocabulary dimension, producing the output logits.
We explored the impact of removing this projection head (w/o lm\_head), where we instead use the hidden states from the last layer of the LLM as features for downstream prediction.
The inferior result in Table~\ref{tab:ab2} demonstrates the vocabulary distribution learned during distillation remains indispensable.
Furthermore, removing the xLSTM head (w/o xLSTM) led to decreased performance, underscoring the critical role of xLSTM in enhancing sequential feature extraction.

\subsubsection{Round of Conversations}
Additionally, the influence of conversation rounds on model performance was investigated, building on the settings discussed in Sec.~\ref{sec:img_desc}, which considers the content of context by various aspects.

Initially, we tested a configuration without fine-tuning the local MLLM with an online MLLM (w/o ft) and without image descriptions (w/o ID), relying solely on the local MLLM with a simple prompt requesting quality results within categories [\textit{bad, poor, fair, good, excellent}], this configuration unsurprisingly yielded the poorest performance.
Then with the image description (w/o ft, w/ ID) performs worse,
even when incorporating a full conversation including the generated final quality result (w/o ft, w/ FullConv), performance decreases further.
This is likely because the original local MLLM did not align well with human perspective evaluation, leading to increased bias with more descriptive context.
By using image descriptions for fine-tuning the local MLLM, but not providing them as context during the downstream training stage (w/ ft, w/o ID), the model outperforms the configurations without fine-tuning, indicating the strong image description capabilities related to the three aspects from online MLLM can be effectively transferred to the local MLLM.
Furthermore, we tested the configuration to infer the classification result (w/ ft, w/ FullConv), the performance decline could be caused by the determinate results potentially misleading the subsequent training step.
Our M3-AGIQA implementation (w/ ft, w/ ID), which leverages conversations including an initial prompt from user, a response from MLLM, and another user prompt for asking for the result, further emphasizes the crucial role of image descriptions in achieving the results outlined in Sec.~\ref{sec:img_desc}.

\subsection{Variant Study - RQ4} \label{sec:variant}

\subsubsection{Effect of Different Pooling Strategies.} We conducted experiments on the pooling methods used after the xLSTM head, as detailed in Fig.~\ref{fig:variants} (a).
Using only the last token (last) for pooling resulted in poor performance, indicating insufficient contextual information was captured.
Implementing max pooling (max) led to even worse outcomes, suggesting a significant loss of context.
Alternatively, using a mean of the first and last tokens (fl mean) also proved inferior to our method with mean pooling, which was found to best preserve rich contextual data.

\subsubsection{Comparison of Sequence Modeling Architectures.}
We substitute the xLSTM head with four representative sequence modeling architectures (LSTM~\cite{hochreiter1997long}, GRU~\cite{cho2014learning}, Transformer~\cite{vaswani2017attention}, and Mamba2~\cite{mamba2}) and a vanilla MLP, ensuring that all sequence models and the MLP are configured with similar parameter budgets.
Results in Fig.~\ref{fig:variants} (b) indicate that the vanilla MLP baseline yields the lowest performance, which highlights the necessity of sequential modeling for this task.
Mamba2 and Transformer configurations outperform LSTM and GRU, consistent with prior findings that these architectures are more effective for capturing long-range dependencies.
Our implementation with the xLSTM head achieves the best results, benefiting from its specially designed modules for sequential feature extraction.

\subsubsection{Local MLLM Backbone Variants.} 
\begin{figure}
    \centering
    \begin{subfigure}{.325\linewidth}
        \includegraphics[width=\textwidth]{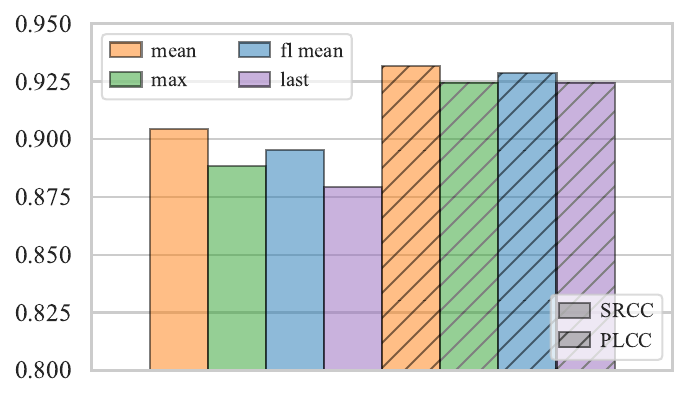}
        \caption{Pooling Strategies}
    \end{subfigure}
    \begin{subfigure}{.325\linewidth}
        \includegraphics[width=\textwidth]{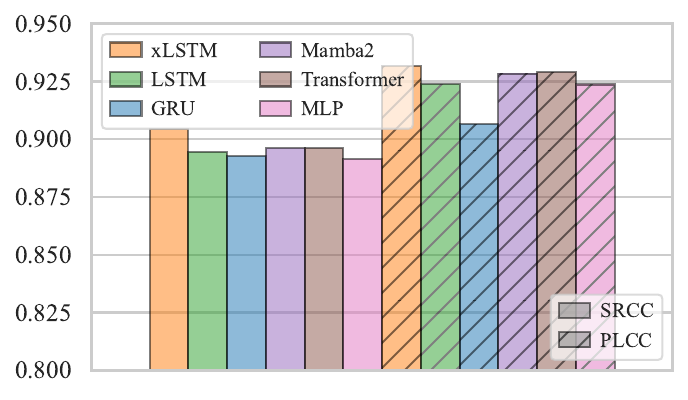}
        \caption{Sequence Model Variants}
    \end{subfigure}
    \begin{subfigure}{.325\linewidth}
        \includegraphics[width=\textwidth]{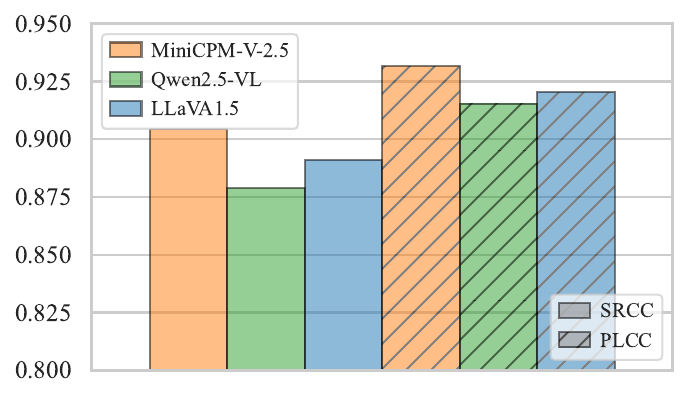}
        \caption{Local MLLM Backbones}
    \end{subfigure}

    \caption{(a) Comparison of different pooling strategies applied after the xLSTM head. ``mean'' refers to mean pooling, ``max'' to max pooling, ``fl mean'' to mean of the first and last token, and ``last'' to last-token pooling. 
(b) Comparison of different sequence modeling architectures used as the feature extractor, including xLSTM, LSTM, GRU, Mamba2, Transformer, and vanilla MLP. 
(c) Comparison of different local MLLMs as the backbone, including MiniCPM-V-2.5, Qwen2.5-VL, and LLaVA1.5.
For all plots, SRCC and PLCC metrics are reported on the AGIQA-3k~\cite{li2023agiqa} dataset (quality aspect). Higher values indicate better performance.}
    \label{fig:variants}
\end{figure}

We further conducted experiments using different local open-source MLLMs as the backbone for our framework, including \textit{openbmb/MiniCPM-Llama3-V-2\_5}~\cite{yao2024minicpmv}, \textit{llava-hf/llava-1.5-7b-hf}~\cite{liu2024improved}, and \textit{Qwen/Qwen2.5-VL-7B-Instruct}~\cite{bai2025qwen2}.
Each model was fine-tuned with the same LoRA settings and evaluated under identical conditions. 
As summarized in Fig.~\ref{fig:variants} (c), MiniCPM-V-2.5 demonstrated the best overall performance on our AGIQA benchmarks, likely due to its stronger multimodal reasoning and compact architecture, which facilitated efficient adaptation. 
In contrast, LLaVA1.5 and Qwen2.5-VL lagged slightly in quality prediction, possibly owing to differences in pretraining data and architecture.
These results suggest that future work should consider pretraining strategies and model selection carefully when designing MLLM-based AGIQA systems.

\begin{table*}
    \centering
    \caption{Cross-dataset validation on quality aspect, results of methods with asterisk symbol ``$^\ast$'' are directly retrieved from corresponding papers.}
    \label{tab:crossdataset1}
    \resizebox{\textwidth}{!}{
    \begin{tabular}{c|cc|cc|cc|cc}
        \toprule
        \multirow{2}{*}{Models} & \multicolumn{2}{c|}{AGIQA-3k$\rightarrow$AIGCIQA2023} & \multicolumn{2}{c|}{AGIQA-3k$\rightarrow$AIGIQA-20k} & \multicolumn{2}{c|}{AIGCIQA2023$\rightarrow$AGIQA-3k} & \multicolumn{2}{c}{AIGIQA-20k$\rightarrow$AGIQA-3k}  \\
        \cline{2-9}
    & SRCC & PLCC & SRCC & PLCC & SRCC & PLCC & SRCC & PLCC \\
        \midrule
        VGG16~\cite{Simonyan15} & 0.6373 & 0.6429 & 0.6874 & 0.7045 & 0.6017 & 0.6396 & 0.7149 & 0.7221 \\ 
        ResNet50~\cite{he2016deep} & 0.6749 & \underline{0.6786} & 0.6695 & 0.7235 & 0.6752 & \underline{0.7475} & 0.6130 & 0.4861 \\
        ViT/B/16~\cite{dosovitskiy2020vit} & 0.5662 & 0.5744 & 0.6659 & 0.7133 & 0.5057 & 0.5579 & 0.7497 & 0.7493 \\
        ViL~\cite{alkin2024vision} & 0.6088 & 0.6033 & 0.6894 & 0.7459 & 0.6574 & 0.6731 & 0.7829 & 0.8090 \\
        \midrule
        DBCNN$^\ast$~\cite{zhang2018blind} & 0.654 & 0.664 & - & - & 0.627 & 0.688 & - & - \\
        StairIQA~\cite{sun2022blind} & 0.6681 & 0.6566 & 0.7001 & 0.7579 & 0.5958 & 0.6264 & 0.7606 & 0.7825 \\
        MGQA~\cite{wang2021multi} & 0.6609 & 0.6696 & 0.6730 & 0.7348 & 0.6090 & 0.6624 & 0.7315 & 0.7363 \\
        HyperIQA~\cite{Su_2020_CVPR} & 0.6451 & 0.6521 & 0.6780 & 0.7115 & 0.4834 & 0.5241 & 0.6252 & 0.6014 \\
        \midrule
        AMFF-Net$^\ast$~\cite{zhou2024adaptive} & 0.678 & 0.669 & - & - & 0.654 & 0.695 & - & - \\
        MA-AGIQA~\cite{wang2024large} & \underline{0.7108} & \underline{0.7014} & \underline{0.7358} & \textbf{0.7998} & 0.6495 & \underline{0.7219} & 0.7967 & \underline{0.8238} \\
        IPCE~\cite{peng2024aigc} & {0.6784} & 0.6558 & \textbf{0.7771} & \underline{0.7727} & \underline{0.7170} & 0.7042 & \underline{0.8231} & 0.8232\\
        \midrule
        M3-AGIQA (Ours) & \textbf{0.7489} & \textbf{0.7461} & 0.6862 & 0.7340 & \textbf{0.7427} & \textbf{0.7542} & \textbf{0.8452} & \textbf{0.8772} \\
        M3-AGIQA (w/ TP) & \textit{0.8557} & \textit{0.8845} & \textit{0.8973} & \textit{0.9308} & \textit{0.9027} & \textit{0.9314} & \textit{0.8961} & \textit{0.9310} \\
        \bottomrule
    \end{tabular}
    }
\end{table*}

\subsection{Cross-dataset Validation - RQ5}

In order to demonstrate the generalization ability of our method, we further conduct cross-dataset validation experiments. 
Following the previous settings, we focused solely on the quality aspect.
For our method, we used the fine-tuned local MLLM from dataset $A$ to generate intermediate image quality descriptions for dataset $B$, and predicted outcomes on $B$ accordingly. 
For baselines, we simply loaded the checkpoint trained on $A$ and predicted on $B$, without using any descriptions.
As shown in Table~\ref{tab:crossdataset1}, the results indicate that our method has a strong generalization ability compared with its counterparts.
While our method demonstrates strong generalization when transferring from larger or similar-sized datasets to smaller ones, its performance in the small-to-large dataset transfer scenario (AGIQA-3k~\cite{li2023agiqa} to AIGIQA-20k~\cite{li2024aigiqa}) is less competitive compared to MA-AGIQA and IPCE. 
This suggests that future work could explore strategies such as more advanced domain adaptation or data augmentation to improve scalability and generalization to substantially larger target datasets.
Also, we tried with additionally training the predictor (w/ TP), the performance was close to that of fully trained on that dataset, which proves the effectiveness of the predictor.

\section{Limitations} \label{sec:limit}
Our work uses LoRA fine-tuning on a $8B$ parameter local MLLM, which is practical on widely available hardware.
Additionally, we relied on an online API for distillation, introducing extra costs when initiating projects from scratch.
Despite these efforts, as observed in Sec.~\ref{sec:img_desc}, the local MLLM struggled to effectively handle all three aspects simultaneously, likely stemming from the inadequate capacity of the vision encoder within the MLLM.
Furthermore, the ranking performance could be enhanced by incorporating specific constraints, such as a ranking loss function, although this would require more computational resources and larger batch sizes.

\section{Ethics Statement} \label{sec:ethics}
Since our method employs an MLLM, which includes a language model for text generation, there is a potential risk of generating harmful content.
However, we rely on the online MLLM's robust safeguards to prevent harmful outputs from normal prompts, significantly mitigating such risks.
In the datasets used, fewer than $50$ images are labeled as Not Safe For Work (NSFW), but these images were deemed acceptable by human judgment and rejected only because of the stringent filtering of the \textit{Gemini} API.
For these cases, \textit{GPT-4o} was leveraged as a safe and effective alternative, ensuring high-quality content generation while preserving the integrity and applicability of the research.
\section{Conclusion} \label{sec:conclusion}
This paper presents M3-AGIQA, a comprehensive framework for assessing AI-generated image quality using a multimodal, multi-round, and multi-aspect strategy.
By distilling capabilities from an online MLLM to a local model, M3-AGIQA aligns closely with human perception in quality, correspondence, and authenticity.
Experimental results demonstrate superior performance compared with state-of-the-art methods on several established benchmarks.
The M3-AGIQA framework provides a promising recipe for practical deployment in content moderation or generative image curation pipelines, though further validation in real-world settings is warranted.
While computational challenges remain, M3-AGIQA sets the stage for future research in efficient AIGI quality assessment for broader generalization and improved computational efficiency.

\bibliographystyle{ACM-Reference-Format}
\bibliography{ref}

\newpage
\appendix 
\definecolor{positive}{RGB}{0,204,0}
\definecolor{negative}{RGB}{234,107,102}

\section{Prompts} \label{app:prompts}
\subsection{Interaction with Online MLLM Service} \label{app:prompt_getid}
This is the prompt template used for retrieving the image descriptions on the three aspects from an online MLLM. The ``alignment'' aspect in the template is equivalent to ``correspondence''. 

\noindent\begin{minipage}{\textwidth}
\begin{mdframed}[linecolor=black,roundcorner=5pt,skipabove=5pt,skipbelow=5pt]
\textbf{Description Prompt Template}

\textbf{User:} Analyze this image generated from the text prompt: `\{\textit{prompt}\}'.

The quality score is \{\textit{mos\_q}\}, alignment score is \{\textit{mos\_a}\}, authenticity score is \{\textit{mos\_au}\}. Scores are between \{\textit{min}\} and \{\textit{max}\}, with higher scores being better.

Provide your assessment in JSON format with the following keys:

quality\_explanation: Describe the overall quality of the image, considering aspects such as composition, clarity, and any noticeable flaws.

alignment\_explanation: Explain how well the image reflects the elements and intent of the provided prompt. Be specific about which aspects of the prompt are successfully conveyed and which might be missing or misinterpreted.

authenticity\_explanation: Discuss how closely the image resembles real artworks. Highlight any parts of the image that appear non-real or artificial.

Do not include the numerical scores in your response.

**Example:**

\{

  ``quality\_explanation'': ``The overall quality of the image is quite high. The composition is balanced, with the boat centered and leading the viewer's eye towards the bridge and beyond. The clarity is excellent, with sharp details on the boat, water, and surrounding cliffs. The lighting is dramatic and enhances the overall atmosphere of the scene. However, there are some noticeable flaws, such as the overly saturated colors, which can detract from the natural feel of the image.'',
  
  ``alignment\_explanation'': ``The image closely aligns with the prompt 'bridge over a body of water with a boat in the water.' The bridge is prominently featured, and the boat is clearly visible in the water. The scene captures the essential elements of the prompt well. However, some additional details, such as the type of bridge or the style of the boat, could have been more specific to better reflect the intent of the prompt.'',
  
  ``authenticity\_explanation'': ``The image has a surreal, almost dreamlike quality, which makes it less authentic as a representation of a real-world scene. The colors are highly saturated and the lighting effects are dramatic, which enhances the artistic feel but reduces the realism. The boat and the bridge look more like artistic renditions rather than actual structures, and the overall scene feels more like a digital artwork or a scene from a fantasy world rather than a photograph of a real place.''
  
\}
\end{mdframed}
\end{minipage}

\subsection{One-round Conversation} \label{app:prompt_oneround}
This is the prompt template for the configuration without the image description (w/o ID) mentioned in our experiments.

\noindent\begin{minipage}{\textwidth}
\begin{mdframed}[linecolor=black,roundcorner=5pt,skipabove=5pt,skipbelow=5pt]
\textbf{One-round Conversation Prompt Template (Qual.)}

\textbf{User:} Take a close look at this AI-generated image and tell me everything you can about what you see. It's original prompt is: `\{\textit{prompt}\}'. Could you please help to rate the image based on it's overall quality? Please give me a result from [\textit{bad, poor, fair, good, excellent}].
Please just output one word from the list.
\end{mdframed}
\end{minipage}

\noindent\begin{minipage}{\textwidth}
\begin{mdframed}[linecolor=black,roundcorner=5pt,skipabove=5pt,skipbelow=5pt]
    \textbf{One-round Conversation Prompt Template (Corr.)}
    
    \textbf{User:} Take a close look at this AI-generated image and tell me how well it aligns with the elements and intent of the original prompt: `\{\textit{prompt}\}'. Could you please help to rate the image based on its correspondence with the prompt? Please give me a result from [\textit{bad, poor, fair, good, excellent}].
    Please just output one word from the list.
    \end{mdframed}
\end{minipage}

\noindent\begin{minipage}{\textwidth}
\begin{mdframed}[linecolor=black,roundcorner=5pt,skipabove=5pt,skipbelow=5pt]
    \textbf{One-round Conversation Prompt Template (Auth.)}
    
    \textbf{User:} Take a close look at this AI-generated image and evaluate how authentic it appears. Does it resemble real artworks or scenes? Highlight any parts that seem artificial or non-realistic. Could you please help to rate the image based on its authenticity? Please give me a result from [\textit{bad, poor, fair, good, excellent}].
    Please just output one word from the list.
    \end{mdframed}
\end{minipage}

\subsection{Multi-round Conversation} \label{app:prompt_multiround}
The prompt template is for fine-tuning the local MLLM with the response from the online MLLM service, and is also used during the inference stage.

\noindent\begin{minipage}{\textwidth}
\begin{mdframed}[linecolor=black,roundcorner=5pt,skipabove=5pt,skipbelow=5pt]
\textbf{Multi-round Conversation Prompt Template (Qual.)}

\textbf{User:} Please closely examine this AI-generated image and provide a detailed analysis of its content and quality. The original prompt for the image was: `\{\textit{prompt}\}' What can you deduce about the image based on this prompt, and how would you assess its overall quality?

\textbf{Assistant:} \{\textit{response\_0}\}

\textbf{User:} Based on your analysis, could you please rate the image's overall quality? Choose one word from the following list: [\textit{bad, poor, fair, good, excellent}], where the words range from low to high quality.

\textbf{Assistant:} \{\textit{response\_1}\}
\end{mdframed}
\end{minipage}

\noindent\begin{minipage}{\textwidth}
\begin{mdframed}[linecolor=black,roundcorner=5pt,skipabove=5pt,skipbelow=5pt]
    \textbf{Multi-round Conversation Prompt Template (Corr.)}
    
    \textbf{User:} Please closely examine this AI-generated image and provide a detailed analysis of how well it aligns with the original prompt: `\{\textit{prompt}\}'.
    
    \textbf{Assistant:} \{\textit{response\_0}\}
    
    \textbf{User:} Based on your analysis, could you please rate the image's alignment with its original prompt? Choose one word from the following list: [\textit{bad, poor, fair, good, excellent}], where the words indicate the degree of alignment from low to high.
    
    \textbf{Assistant:} \{\textit{response\_1}\}
    \end{mdframed}
\end{minipage}

\noindent\begin{minipage}{\textwidth}
\begin{mdframed}[linecolor=black,roundcorner=5pt,skipabove=5pt,skipbelow=5pt]
    \textbf{Multi-round Conversation Prompt Template (Auth.)}
    
    \textbf{User:} Please closely examine this AI-generated image and provide a detailed analysis of its authenticity. The original prompt for the image was: `\{\textit{prompt}\}'. How closely does the image resemble real artworks or scenes? Highlight any parts of the image that appear non-real or artificial.
    
    \textbf{Assistant:} \{\textit{response\_0}\}
    
    \textbf{User:} Based on your analysis, could you please rate the image's authenticity? Choose one word from the following list: [\textit{bad, poor, fair, good, excellent}], where the words indicate the degree of authenticity from low to high.
    
    \textbf{Assistant:} \{\textit{response\_1}\}
    \end{mdframed}
\end{minipage}

Notice that during fine-tuning (Init Description stage), the \textit{repsonse\_0} is the response from online MLLM service and \textit{response\_1} is the ground truth by categorize the MOS to the five string labels.
While during the Inference stage, both \textit{response\_0} and \textit{response\_1} are generated by the fine-tuned local MLLM.

\section{Cases}
\subsection{Multi-Aspect Responses} \label{app:responses}
To illustrate the effectiveness of the distillation, we selected two images generated using the same prompt but exhibiting different quality levels, as shown in Table~\ref{tab:multiaspect}.
The responses from both the online MLLM service and the fine-tuned local MLLMs were analyzed. 
Highlighted phrases such as ``high quality'', ``perfectly aligns'', and ``does not resemble real artworks'' in the table demonstrate that the fine-tuned local MLLMs successfully learned to evaluate the images across multiple aspects, reflecting the mechanism transferred from the online MLLM service.
\begin{table*}
    \centering
    \caption{Multi-Aspect responses from online MLLM service and fine-tuned local MLLM for the image generated by prompt \textit{an aerial photo of a sandy island in the ocean}}\label{tab:multiaspect}
    \resizebox{1.0\textwidth}{!}{
    \begin{tabular}{c m{0.8cm} m{5.5cm}m{5.5cm}}
        \toprule
        Image & Aspect & ID from online MLLM service & ID from fine-tuned local MLLM \\
        \midrule 
        \multirow{3}{*}{
        \begin{minipage}{.2\textwidth}
            \includegraphics[width=\linewidth]{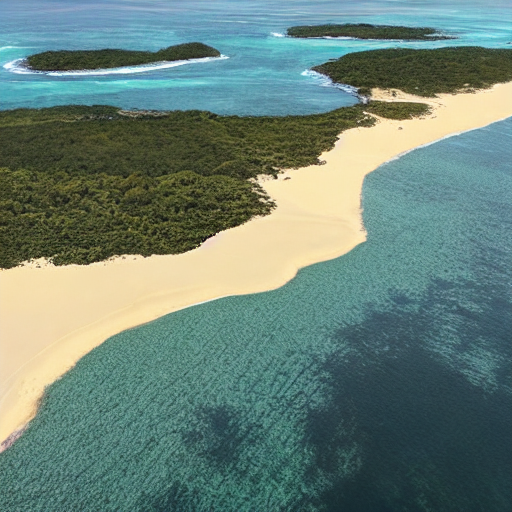}
        \end{minipage}} &
        Qual. & The image boasts \textcolor{positive}{\textbf{high quality with a clear, well-composed aerial view}}. The colors are \textcolor{positive}{\textbf{vibrant}}, and the textures of the sand, water, and vegetation are \textcolor{positive}{\textbf{well-defined}}. There are \textcolor{positive}{\textbf{no noticeable flaws or artifacts}}, contributing to a realistic and \textcolor{positive}{\textbf{visually appealing}} image. &
        The image boasts \textcolor{positive}{\textbf{excellent clarity and composition}}. The aerial perspective is well-executed, providing a \textcolor{positive}{\textbf{clear view}} of the island's shape and the surrounding waters. The lighting is \textcolor{positive}{\textbf{natural and flattering}}, highlighting the textures of the sand and foliage. Overall, the image is of \textcolor{positive}{\textbf{high quality}}.
        \\
        \cline{2-4}
        & Corr. & The image \textcolor{positive}{\textbf{aligns perfectly}} with the prompt. It depicts a sandy island, partially covered in vegetation, situated in a vast expanse of blue ocean. The aerial perspective \textcolor{positive}{\textbf{accurately reflects}} the ``aerial photo'' aspect of the prompt. The image successfully encapsulates \textcolor{positive}{\textbf{all elements}} of the prompt. &
        The image \textcolor{positive}{\textbf{perfectly aligns}} with the prompt. It \textcolor{positive}{\textbf{clearly depicts}} a sandy island in the ocean from an aerial perspective. \textcolor{positive}{\textbf{All elements}} of the prompt are present and \textcolor{positive}{\textbf{accurately represented}}. \\
        \cline{2-4}
        & Auth. & The image \textcolor{positive}{\textbf{closely resembles a real aerial photograph}}. The colors, while vibrant, remain within a \textcolor{positive}{\textbf{natural palette}}, and the lighting and shadows appear realistic. The textures and patterns of the sand, water, and vegetation contribute to the image's authenticity. There are \textcolor{positive}{\textbf{no elements that suggest artificiality or manipulation}}. & 
        The image \textcolor{positive}{\textbf{closely resembles a real aerial photograph}} of a tropical island. The textures of the sand, water, and vegetation appear \textcolor{positive}{\textbf{natural and realistic}}. There are \textcolor{positive}{\textbf{no obvious signs of digital manipulation or artificial elements that would detract from its authenticity}}. \\
        \midrule
        \multirow{3}{*}{\begin{minipage}{.2\textwidth}
            \includegraphics[width=\linewidth]{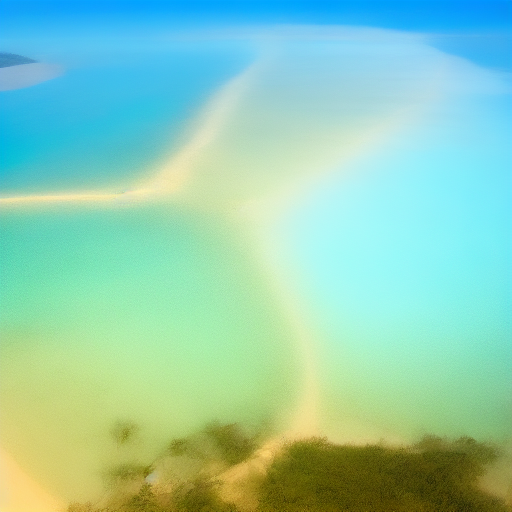}
        \end{minipage}} &
        Qual. & The image quality is \textcolor{negative}{\textbf{relatively low}}. The composition is \textcolor{negative}{\textbf{poorly}} defined, lacking a clear focal point.  While the colors are vibrant, they are \textcolor{negative}{\textbf{heavily blurred and lack detail}}, making the overall image appear \textcolor{negative}{\textbf{hazy and unrealistic}}. & The overall quality of the image is \textcolor{negative}{\textbf{quite low}}. While it attempts to depict an aerial view of a sandy island in the ocean, it \textcolor{negative}{\textbf{lacks clarity and detail}}. The colors are overly saturated and blend together, creating a \textcolor{negative}{\textbf{hazy, unrealistic effect}}. The shapes of the landforms are \textcolor{negative}{\textbf{indistinct and lack definition}}. \\
        \cline{2-4}
        & Corr. & The image \textbf{partially reflects the prompt}. It features a sandy area surrounded by blue, which could be interpreted as an island in the ocean from an aerial perspective. However, the \textcolor{negative}{\textbf{layout and shapes are ambiguous}}, and it's difficult to definitively identify the scene as an island. The \textcolor{negative}{\textbf{lack of clarity}} hinders a stronger alignment with the prompt. & 
        The image \textbf{partially aligns with the prompt}. It does depict a body of water and what could be interpreted as a sandy island in the distance. However, the aerial perspective is \textcolor{negative}{\textbf{not clearly conveyed}}, and the'sandy island' appears more as a \textcolor{negative}{\textbf{vague shape}} rather than a defined landmass. The \textcolor{negative}{\textbf{lack of detail}} makes it difficult to confirm the presence of sand. \\
        \cline{2-4}
        & Auth. & The image \textcolor{negative}{\textbf{does not resemble real artwork or a photograph}}. The excessive \textbf{blurring, unrealistic} color gradients, and \textcolor{negative}{\textbf{lack of detail}} give it a \textcolor{negative}{\textbf{heavily artificial and digitally manipulated feel}}. It lacks the characteristics of a genuine aerial photograph. &
        The image \textcolor{negative}{\textbf{does not resemble real artworks or photographs}}. The excessive \textcolor{negative}{\textbf{blurring, unrealistic color palette}}, and \textbf{lack of detail} contribute to an \textcolor{negative}{\textbf{artificial and almost abstract appearance}}. It \textcolor{negative}{\textbf{lacks the realism}} and texture expected in a depiction of a natural landscape. \\
        \bottomrule
    \end{tabular}}
\end{table*}
\subsection{Good and Bad Cases} \label{app:goodnbadcases}
\begin{table*}
    \centering
    \caption{Examples showcasing the top three best-performing cases and the bottom three worst-performing cases on dataset AGIQA-3k~\cite{li2023agiqa} quality aspect.}\label{tab:badngoodcases}
    \resizebox{1.0\textwidth}{!}{
    \begin{tabular}{m{2cm}c m{5.5cm}m{5.5cm}cc}
        \toprule
        Prompt &
        Image & ID from online MLLM service & ID from fine-tuned local MLLM & label & predicted \\
        \midrule
        photograph of beautiful landscape, baroque style &
        \begin{minipage}{.2\textwidth}
            \includegraphics[width=\linewidth]{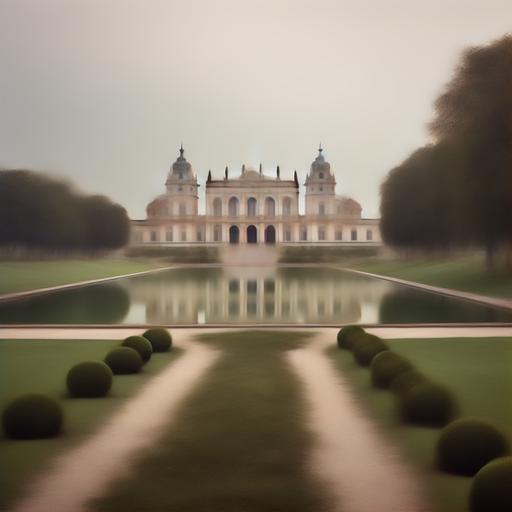}
        \end{minipage} & 
        The image quality is \textcolor{positive}{\textbf{good}}, featuring a \textcolor{positive}{\textbf{clear and well-composed}} scene. The building is sharply focused, and the reflection in the water is \textcolor{positive}{\textbf{well-rendered}}. The soft, diffused lighting contributes to a serene atmosphere. However, the overall image appears \textcolor{negative}{\textbf{somewhat blurry, lacking fine detail}}, particularly in the foliage and the foreground. & 
        The overall quality of the image is \textcolor{positive}{\textbf{good}}. The composition is \textcolor{positive}{\textbf{well-balanced}}, with the grandeur of the baroque-style building drawing the viewer's eye. The use of a symmetrical path leading to the building creates a sense of depth and perspective. However, the image is \textcolor{negative}{\textbf{somewhat blurry}}, particularly in the background, which detracts from the sharpness and detail. &
        3.0 & 3.0015\\

        \midrule
        an emo portrait painting &
        \begin{minipage}{.2\textwidth}
            \includegraphics[width=\linewidth]{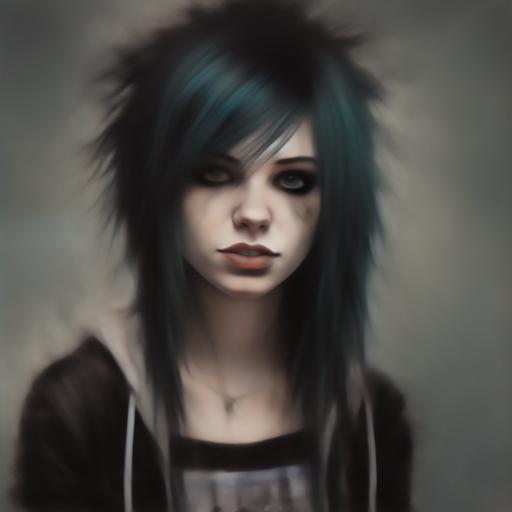}
        \end{minipage} & 
        The image boasts \textcolor{positive}{\textbf{high quality}}, especially in its technical execution. The portrait showcases \textcolor{positive}{\textbf{smooth blending}}, creating a \textcolor{positive}{\textbf{realistic}} portrayal of skin and hair. The use of light and shadow adds depth and dimension to the face. However, the \textcolor{negative}{\textbf{background lacks detail}}, appearing \textcolor{negative}{\textbf{blurry and undefined}}, which slightly detracts from the overall composition. &
        The image quality is \textcolor{positive}{\textbf{good}} for a digital painting. The brushstrokes are visible and give a sense of texture to the hair and skin. The lighting is \textcolor{positive}{\textbf{well-executed}}, creating a moody atmosphere. However, the face appears \textcolor{negative}{\textbf{slightly blurry}}, particularly around the eyes and mouth, which detracts from the overall clarity. &
        2.9417 & 2.9434 \\
        
        \midrule
        a masonic illuminati waiting room with a couch, soft lighting, abstract style &
        \begin{minipage}{.2\textwidth}
            \includegraphics[width=\linewidth]{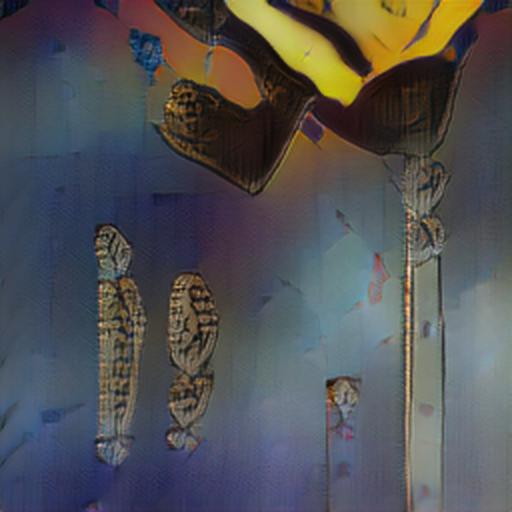}
        \end{minipage} & 
        The image is of \textcolor{negative}{\textbf{low quality}}, appearing more as an abstract composition than a coherent scene. It \textcolor{negative}{\textbf{lacks clarity}}, with \textcolor{negative}{\textbf{blurry and indistinct}} elements. The lighting, while potentially aiming for `soft,' is \textcolor{negative}{\textbf{muddled}} and doesn't contribute to a cohesive atmosphere. & 
        The image is of \textcolor{negative}{\textbf{low quality}}. It is \textcolor{negative}{\textbf{blurry}}, \textcolor{negative}{\textbf{lacks clear shapes}}, and the colors are \textcolor{negative}{\textbf{muddled}}. There is no discernible composition, and the overall impression is abstract and unclear. & 
        0.9708 & 0.9736 \\
        
        \midrule
        the large cresting waves & 
        \begin{minipage}{.2\textwidth}
            \includegraphics[width=\linewidth]{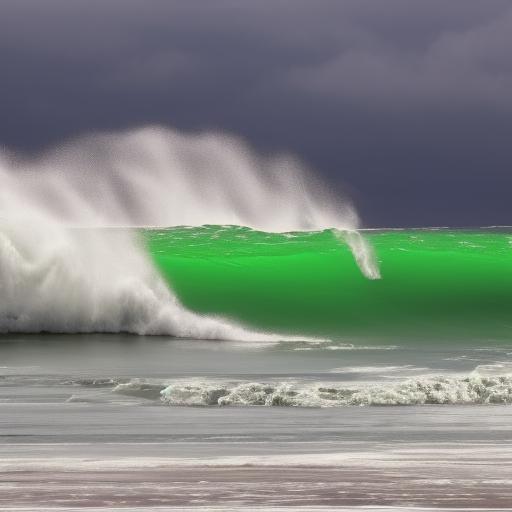}
        \end{minipage} & 
        The image quality is low due to the \textcolor{negative}{\textbf{unrealistic}}, \textcolor{negative}{\textbf{almost neon green color}} of the water. The composition is simple, focusing solely on the wave, and \textcolor{negative}{\textbf{lacks depth}}. The spray of the wave is well-detailed, but the overall image feels \textcolor{negative}{\textbf{artificial}}. & 
        The image quality is \textcolor{positive}{\textbf{decent}} but not exceptional. The wave itself is well-rendered with a \textcolor{positive}{\textbf{good sense of movement and power}}. However, the composition is somewhat \textcolor{negative}{\textbf{simplistic}}, focusing solely on the wave without much context or background detail. The color palette is \textcolor{negative}{\textbf{limited}}, with the green of the wave standing out against the gray sky and water. &
        1.6788 & 2.7090\\

        \midrule
        a parade of disconnected images : a tragic supernova, hyper detail, HDR lighting & 
        \begin{minipage}{.2\textwidth}
            \includegraphics[width=\linewidth]{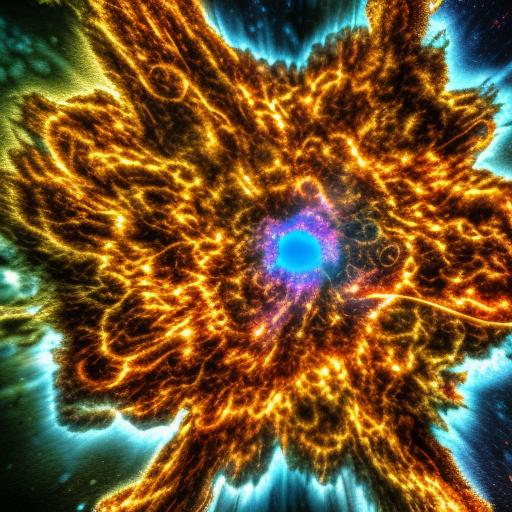}
        \end{minipage} & 
        The image is of \textcolor{positive}{\textbf{high quality}} in terms of clarity and detail. The colors are \textcolor{positive}{\textbf{vibrant}} and the textures are \textcolor{positive}{\textbf{intricate}}. The overall composition, however, \textcolor{negative}{\textbf{lacks a clear focal point}} and can feel overwhelming due to the lack of negative space. &
        The image boasts \textcolor{positive}{\textbf{high quality}} in terms of clarity and detail. The textures are intricate, and the lighting effects are \textcolor{positive}{\textbf{well-executed}}, creating a sense of depth and dynamism. The composition, however, is somewhat lacking. While the image is visually striking, it \textcolor{negative}{\textbf{lacks a clear focal point}} and feels somewhat \textcolor{negative}{\textbf{chaotic}}. & 
        1.2288 & 2.5059 \\

        \midrule
        a very drunk person named aaron, sci-fi style & 
        \begin{minipage}{.2\textwidth}
            \includegraphics[width=\linewidth]{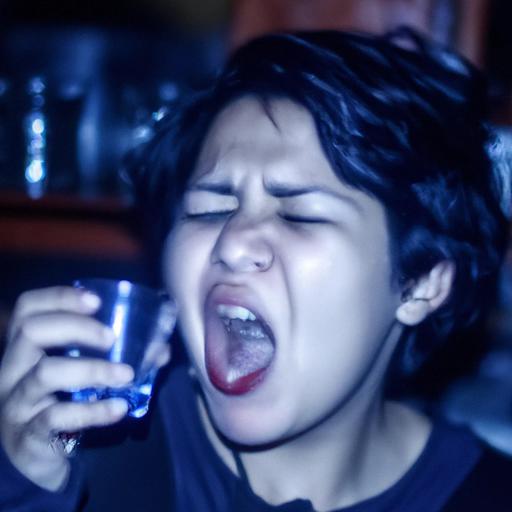}
        \end{minipage} & 
        The image quality is \textcolor{negative}{\textbf{quite poor}}. It appears to be a poorly lit photograph with a strong blue color cast, making it \textcolor{negative}{\textbf{difficult to discern details}}. The image is \textcolor{negative}{\textbf{blurry}}, particularly in the background, further reducing the overall clarity. & 
        The image quality is \textcolor{negative}{\textbf{quite low}}. It appears to be a \textcolor{negative}{\textbf{blurry}}, poorly lit photograph rather than a well-composed image. The lighting is \textcolor{negative}{\textbf{harsh and uneven}}, creating an unflattering effect on the subject's face. &
        3.7735 & 2.4688\\
        \bottomrule
    \end{tabular}}
\end{table*}

We analyzed the cases where our method's predictions were most and least aligned with the ground truth by sampling the top three aligned and the bottom three misaligned instances as illustrated in Table~\ref{tab:badngoodcases}. The well-aligned cases shared similar tones in their descriptions, indicating close agreement between the predicted and actual image quality assessments.

Conversely, among the misaligned examples, the fourth case illustrates a significant discrepancy. The ground truth description labeled the image as having an ``unrealistic, almost neon green color'' and an ``artificial'' feel, indicating poor quality. However, the predicted description portrayed the image more positively, noting ``decent'' quality with a ``sense of movement and power.'' This led to a prediction of ``good'' ($2.7090$) in contrast to the actual ``poor'' ($1.6788$) rating, highlighting a fundamental difference in interpretation that affected the accuracy of our model's prediction.

In the fifth case, the ground truth rated the image as ``poor'' ($1.7111$), yet the description provided by the online MLLM service and the fine-tuned local MLLM seems to be more positive with terms like ``high quality'' and ``vibrant''.
As a result, the prediction is ``fair'' ($2.5059$), which misaligned with the ground truth but aligned well with the overly positive image descriptions provided.
This case testifies the importance for quality of image descriptions on the other side, if the image descriptions used for fine-tuning is more precisely aligned with the human perceptual evaluation of ``poor'' ($1.2288$), the prediction would have been better.
In this case, a better description could be like:

\noindent\begin{minipage}{\textwidth}
\begin{mdframed}[linecolor=black,roundcorner=5pt,skipabove=5pt,skipbelow=5pt]
This image's low rating likely stems from its overwhelming brightness and chaotic composition, which obscures intended details and focus. Although the prompt calls for a ``tragic supernova'' with ``hyper detail'' and ``HDR lighting'', the result lacks a clear structure or focal point, making interpretation challenging. Intense yellow and blue hues dominate without balance, creating visual confusion rather than conveying the intended sense of tragedy.
\end{mdframed}
\end{minipage}

And the last bad case looks just on the contrary to the fifth case, the image descriptions from both online MLLM service and fine-tuned local MLLM are all negative which can be concluded as ``poor'', while the ground truth is ``good'' ($3.7735$), and the predicted score ``fair'' ($2.4688$) is intended to be aligned with the ground truth. A more fitting description considering the actual quality of the image would be:

\noindent\begin{minipage}{\textwidth}
\begin{mdframed}[linecolor=black,roundcorner=5pt,skipabove=5pt,skipbelow=5pt]
The image quality is good, mainly due to the effective depiction of a "very drunk person" in a stylized, sci-fi setting that captures the mood well. The blurred, exaggerated face suggests intoxication, while blue-tinged sci-fi lighting adds a unique, futuristic touch. However, limited clarity in facial features and the background reduces sharpness and makes it harder to connect with the subject. Overall, the expressive emotion and sci-fi style give the image a distinctive character.
\end{mdframed}
\end{minipage}

\end{document}